\documentclass[12pt]{article}
\pdfoutput=1 
\usepackage{amsmath}
\usepackage{enumerate}
\usepackage[round]{natbib}
\usepackage{amsthm} 
\usepackage{amssymb}
\usepackage{subcaption}

\newcommand{\blind}{0}

\usepackage{rotating}
\usepackage{graphicx}
\usepackage{subcaption}

\usepackage{float}

\usepackage{amsthm,amsmath}

\usepackage{amssymb}
\usepackage{subcaption}
\usepackage{soul, xcolor}

\newtheorem{remark}{Remark}

\usepackage{algorithm}
\usepackage{algpseudocode}

\newcommand{\be}{\begin{equation}\begin{aligned} }
\newcommand{\ee}{\end{aligned}\end{equation}}
\newcommand{\bb}[1]{\mathbb{#1}}
\newcommand{\mc}[1]{\mathcal{#1}}

\DeclareMathOperator{\No}{No}

\newcommand{\bzero}{\textbf{0}}

\newcommand{\ba}{\textbf{a}}

\newcommand{\bbb}{\textbf{b}}

\newcommand{\bG}{\textbf{G}}

\newcommand{\bI}{\textbf{I}}

\newcommand{\bQ}{\textbf{Q}}

\newcommand{\bw}{\textbf{w}}
\newcommand{\bW}{\textbf{W}}
\newcommand{\bx}{\textbf{x}}
\newcommand{\bX}{\textbf{X}}
\newcommand{\by}{\textbf{y}}

\newcommand{\btheta}{ \mbox{\boldmath $ \theta $} }

\newcommand{\bTheta}{ \mbox{\boldmath $\Theta$} }

\date{}

\begin{document}

\def\spacingset#1{\renewcommand{\baselinestretch}%
{#1}\small\normalsize} \spacingset{1}


\if0\blind
{
  \title{\bf Mixed-Stationary Gaussian Process \\
    for Flexible Non-Stationary Modeling \\
    of Spatial Outcomes}
  \author{Leo L. Duan\\
   Duke University\\
    and\\
           Xia Wang\\
    University of Cincinnati\\
        and\\
    Rhonda D. Szczesniak \\
    Cincinnati Children's Hospital Medical Center
    	}
  \maketitle
 \fi

\if1\blind
{
  \bigskip
  \bigskip
  \bigskip
  \begin{center}
    {\LARGE\bf Mixed-Stationary Gaussian Process}
\end{center}
  \medskip
} \fi

\begin{abstract}

Gaussian processes (GPs) are commonplace in spatial statistics. Although many non-stationary models have been developed, there is arguably a lack of flexibility compared to equipping each location with its own parameters. However, the latter suffers from intractable computation and can lead to overfitting. Taking the instantaneous stationarity idea, we construct a non-stationary GP {with the stationarity parameter individually set at each location}. Then we utilize the non-parametric mixture model to reduce the effective number of unique parameters. Different from a simple mixture of independent GPs, the {\it mixture in stationarity} allows the components to be spatial correlated, leading to improved prediction efficiency. Theoretical properties are examined and a linearly scalable algorithm is provided.  The application is shown through several simulated scenarios as well as the massive spatiotemporally correlated temperature data.
\end{abstract}

\noindent%
\vfill

\newpage
\spacingset{1.45} 

\section{Introduction}

Gaussian processes (GPs) have become the standard tool for {spatiotemporal modeling}. Using locations as input to generate covariance, GPs model the spatial outcome as a multivariate Gaussian random variable. {The flexibility of covariance is crucial to modeling based on GPs, especially when non-stationarity is present.} There is an admittedly large literature for non-stationary GPs. Among the well known early work, \cite{higdon1998process} used kernel convolution of a stationary process to generate non-stationarity; \cite{schmidt2003bayesian2} used non-uniform shifts on the spatial location to deform a stationary GP, creating locally varying patterns; low-rank representation of covariance such as the one in \cite{banerjee2008gaussian}  also implicitly induces non-stationarity.

While it is straightforward to \emph{generate} non-stationarity, it is questionable if those models are flexible enough. Specifically, how the stationarity changes from one region to another.  This is important for using GPs for spatial prediction. Indeed, over-estimation of the spatial autocorrelation leads to over-smoothing and under-quantification of the prediction uncertainty; likewise, under-estimation causes the interpolator to have overly large variance and low efficiency.

The pioneering idea that formally addressed this issue was called ``evolutionary spectrum''  in time series study \citep{priestley1965evolutionary}, where a non-stationary process is viewed as an instantaneously stationary spectrum  evolving over time. Later, a similar idea was proposed in spatial modeling \citep{dahlhaus2000likelihood}, except that local regions in the location space replace the instant temporal points. More recently, \cite{gramacy2012bayesian} and \cite{guinness2015likelihood} proposed  models that partition the observed location space into small regions, wherein each partition was separately modeled using a stationary GP.

There are two key {limitations} that motivate this work. The first one is the estimation problem. The instantaneously or piece-wise stationary models often require too many parameters; the large parameters-to-data ratio causes high variance in the parameter estimate, unless one artificially limits the number of stationary regions. To solve this problem, we use the non-parametric {Bayesian} approach of mixture models for those parameters, thereby yielding a small pool of candidate parameters in the posterior. Two regions can share the same parameters, without the need to be contiguous. This borrowing of information substantially reduces the number of parameters needed.

The second issue is fundamentally related to prediction. One commonly overlooked fact is that the simple piecewise structure has zero correlation between any two regions, which is problematic as one can only use part of data for interpolation. {Although one could smooth over the boundaries using model averaging \citep{rasmussen2002infinite} or marginalization \citep{gelfand2005bayesian}, we show that the resulting predictor is not as efficient as the one including a Gaussian covariance directly.  Our proposed non-stationary model can be equivalently viewed as a mixture of dependent GPs, which has a dense-matrix Gaussian correlation given the component assignment. This allows one to simply cluster the spatial data while maintaining the prediction efficiency.}

{The paper is structured as follows. Section \ref{model} constructs the Mixed-Stationary Gaussian processes model (MSGP) and discusses its properties.  Section \ref{estimation} describes an efficient estimation procedure for posterior sampling.  Section \ref{iGP} compares MSGP with the commonly used mixture of independent GPs.  Section \ref{application} shows results from several simulated scenario and an application to the massive spatiotemporal  data from the  North American Regional Climate Change Assessment Program (NARCCAP).  Section \ref{discussion} concludes with several interesting directions for future research. }

\section{Mixed-Stationary Gaussian Processes} \label{model}

\subsection{Basic Notation}

{Suppose we have data collected on $n$ locations, $\bx_1,\ldots,\bx_n$, with {$\bx_{i}$} $\in \mc X \subset \bb R^d$. Let $y_{i}$ be the observed outcome. For exposition {simplicity}, we restrict our attention to continuous univariate $y_{i}$$\in \bb R$. To directly associate the location to an observation, we will use notation $y_i=y(\bx_i)$ intermittently.} Let

\begin{equation*}
  {y(\cdot) \sim GP(\mu(\cdot), K(\cdot, \cdot))}
\end{equation*}
 denote a Gaussian process with the mean function $\mu: \mathbb{R}^d\rightarrow  \mathbb{R}$ and the covariance function $K:\mathbb{R}^d\times \mathbb{R}^d \rightarrow  \mathbb{R}$. This means for a finite-element collection $\{(\bx_1,y_1), \ldots, (\bx_n,y_n) \}$, $y_{1:n}=(y_{1},\ldots, y_n)^{\prime}$ is an $n$-variate Gaussian with
\begin{equation*}
	\begin{aligned}
    \mathbb{E} (y_i) = \mu(\bx_i), \quad \mbox{cov} (y_i,y_{i'}) = K(\bx_i, \bx_{i'}).
	\end{aligned}
\end{equation*}
We assume $\mu(\bx_{i})=0$ for $i=1, \ldots, n$ unless otherwise specified.

\subsection{Non-Stationary Gaussian Process by Mixing Stationarity Parameters}
We now parameterize {the covariance function} $K(\cdot, \cdot)$.  Inspired by the ``instantaneous stationarity'' idea \citep{priestley1965evolutionary},  we first assume that each location $\bx_i$ is equipped with a set of parameters $\btheta_i$. Then to control the number of effective parameters, we utilize the non-parametric Bayesian framework to assign infinite mixture distributions on $\btheta$:
\be
\label{np_model}
   {y(\cdot)} &{\sim GP(\mu(\cdot), K(\cdot, \cdot))},\\
   K(\bx_i,\bx_{i'}; \btheta_i,\btheta_{i'}) & = \int_{\mathbb{ R}^d} \exp(j (\bx_i-\bx_{i'})^{\rm T} \bw) g^{1/2}(\bw; \btheta_i)g^{1/2}(\bw; \btheta_{i'}) d\bw + \sigma^21(i=i'),\\
\pi(\btheta_i) & = \sum_{k=1}^{\infty} p_k \delta_{\btheta^*_k}(\btheta),\\
  p_k &>0, \quad \sum_{k=1}^{\infty} p_k =1, \quad \{p_k\}_{k=1}^{\infty} \sim\pi_0(\{p_k\}_{k=1}^{\infty}), \quad \btheta^*_k \sim \pi_0(\btheta^*_k),
\ee
where $\delta_{\btheta^*}(.)$ is the Dirac measure at point $\btheta^*$, $p_k$'s are the mixture weights corresponding to components assignments, $\pi_0$ denotes the prior, $\sigma^2$ is the variance of independent measurement error; $g$ is the Fourier transform of a stationary covariance function {$\tilde K(\bx_{i}-\bx_{i^{\prime}};\btheta)$}.  It is defined as $g(\bw;\btheta) = \int_{\mathbb{ R}^d} \exp(-j \bx^{\rm T} \bw) \tilde K(\Delta; \btheta)d \Delta$, with {$\Delta=\bx_{i}-\bx_{i^{\prime}}$} and $j=\sqrt{-1}$ being the imaginary unit.

Although infinite mixture is assumed, it is well known that $\btheta_i$ will have a discrete distribution  almost surely in the posterior \citep{james2003simple}. This partitions $\btheta_i$'s into several groups with unique values of $\btheta^*_k$,  $\bTheta^{*}= \{ \btheta^*_k: \sum_{i=1}^n 1(\btheta_i=\btheta^*_k)>0\}$. In this work we consider a Dirichlet {process} on $\btheta_i$ with the concentration parameter $\alpha$ for $\pi_0( \{p_k\}_{k=1}^\infty)$, in stick-breaking representation:
\begin{equation*}
    p_1 = v_1, \quad p_k = v_k\prod_{k'=1}^{k-1} (1-v_{k'}) \quad \text{ for }{k\ge 2}, \quad
    v_k \sim \text{Beta}(1,\alpha).
\end{equation*}
With $\alpha<1$, one obtains the number of groups $|\bTheta^{*}| \ll n$. This can be viewed as ``clustering'' on the parameters.

Notice that we recover the stationary covariance
\begin{equation*}
	\begin{aligned}
    K(\bx_i,\bx_{i'}; \btheta_i,\btheta_{i'}) & = \tilde K(\bx_i-\bx_{i'};\btheta_i) + \sigma^2 1(i=i') \quad \text{ if } \btheta_i = \btheta_{i'}.
	\end{aligned}
\end{equation*}
Therefore, we obtain a GP {by mixing parameters from stationary covariances}, and we call this model Mixed-Stationary GP (MSGP).

Alternatively, this model can be viewed as a mixture of several {\em dependent} GPs:
\be
\label{mix_of_dep_gaus}
    y(\bx_i) & = \mu(\bx_i) + f(\bx_i)+ \epsilon_i,\\
    \pi\big(f(\bx_i)\big) & = \sum_{k=1}^{\infty} p_k \delta_{f^*(\bx_i; \btheta^*_k)}\big(f(\bx_i)\big) ,\\
    {f^*(\cdot; \btheta^*_k)} & {\sim \text{GP}[0, \tilde K(\cdot, \cdot;\btheta^*_k)],}\\
   \mbox{cov}[ f^*(\bx_i; \btheta^*_k), f^*(\bx_{i'}; \btheta^*_{k'}) ] &= K(\bx_i,\bx_{i'}; \btheta^*_k,\btheta^*_{k'}), \qquad  \text{ for } k\neq k',
\ee
{where $i, i^{\prime}\in \{ 1, \ldots, n\}^2$.} The dependence allows covariance to be non-zero across components.

\subsection{Properties}

We present several useful properties of the MSGP.

First, it is obvious that conditioning on $\btheta$, {$y(\bx_i|\btheta_i)$}  in \eqref{np_model} formes a valid Gaussian process on the location space of $\mathcal{X}$. The question is if the marginal {$y(\bx)$} is also a valid stochastic process? We show that this is true.

Let $\{\Omega, \mathcal{F}, P\}$ be a probability space and $\mc X$ be an independent space for locations. For a finite set of points $\bx_1,\ldots, \bx_n\in \mc X$, let $\mc L(y_{1:n};\btheta_{1:n})$ denote the multivariate Gaussian likelihood, $\pi(\btheta_i =\btheta^*_k) =p_k$ is the probability for assigning the value {$\btheta^*_k$ to $\btheta_i$}. Then we have
measure associated with \eqref{np_model} for measurable sets $F_i\in
 \mathcal{F}$ for $i=1,\ldots,n$. We have
\begin{equation}
\label{existence}
	\begin{aligned}
    & \Pi_{\{\bx_{1},\ldots,\bx_{n}\}}(F_1 \times \ldots\times F_n) \\
=& \int_{F_n}\ldots
\int_{F_1}\bigg\{\sum_{k_n\in\{1,\ldots,\infty\}}\ldots\sum_{k_1\in\{1,\ldots,\infty\}} \big [\mc L(y_{1:n};\btheta_{1:n}) \prod_i \pi(\btheta_i =\btheta^*_{k_i})\big]  \bigg\} dy_1 dy_2\ldots dy_n.
	\end{aligned}
\end{equation}

\begin{remark} (Existence of Stochastic Process)
The measure in Equation \ref{existence} satisfies the Komolgorov existence criteria.
 \begin{enumerate}
    \item For all permutations of $\{1,\ldots,n\}$, $\{c_1,\ldots,c_n\}$
        $$\Pi_{\{\bx_{c_1},\ldots,\bx_{c_n}\}}(F_{c_1} \times \ldots\times F_{c_n})
        = \Pi_{\{\bx_{1},\ldots,\bx_{n}\}}(F_{1} \times \ldots\times F_{n}).$$
    \item For all measurable set $F_i\in \bb R$
       $$\Pi_{\{\bx_{1},\ldots,\bx_{n}\}}(F_{1} \times \ldots\times F_{n})=
       \Pi_{\{\bx_{1},\ldots,\bx_{n},\bx_{n+1}\}}(F_{1} \times \ldots\times F_{n} \times \mathbb{R}).
       $$
  \end{enumerate}
\end{remark}

Second, it is useful to understand the behavior of this model {a priori}. We compute the first two moments of {$y_{i}$} after marginalizing out $\btheta_i$ over its prior.

\begin{remark}
(First Two Moments a Priori)
The first and second moments of {$y_i$} a priori are
\be
\bb E_0 (y_i) & = \mu(\bx_i), \\
\text{var}_0 (y_i) & = \sum_{k=1}^\infty p_k \tilde K(0;\btheta^*_k) + \sigma^2,\\
\text{cov}_0 (y_i,{y_{i'}}) & = \sum_{{k_2}=1}^\infty \sum_{{k_1}=1}^\infty p_{k_1} p_{k_2} K(\bx_i,\bx_{i'};\btheta^*_{k_1},\btheta^*_{k_2}){,\:\:\:i\ne i^{\prime}}.
\ee
\end{remark}
One can verify all terms are shift-invariant in $(\bx_i,\bx_i')$. Therefore, this non-stationary process is centered around a stationary process {\em a priori}.

Third, conditioned on $\btheta$, the predictive distribution is Gaussian following conventional Kriging formula \citep{cressie1988spatial}. We now present its marginal form:

\begin{remark} (Prediction Moments)
For unobserved $y^\dagger_i$ at location $\bx^\dagger_i$ for $i=1,\ldots,m$, the predictive distribution is a mixture of Gaussians with the first two moments as:

\begin{equation*}
  \begin{aligned}
1. \:\:\: & \bb E (y^\dagger_i \mid y_{1:n}, \btheta_{1:n})  = \mu(\bx^\dagger_i)
\\ & + \sum_{k=1}^{\infty} p_k K (\bx^\dagger_i,\bx_{1:n};\btheta^*_k, \btheta_{1:n})K^{-1}(\bx_{1:n},\bx_{1:n}; \btheta_{1:n}, \btheta_{1:n})\big[y_{1:n}-\mu(\bx_{1:n})\big], \\
2. \:\:\: & \mbox{var} (y^\dagger_i \mid y_{1:n}, \btheta_{1:n})  =  \sum_{{k}=1}^\infty p_{k} \bigg[
K( \bx^\dagger_i,\bx^\dagger_i;\btheta^*_{k},\btheta^*_{k})
\\
&-K( \bx^\dagger_i,\bx_{1:n};\btheta^*_{k}, \btheta_{1:n}) K^{-1}(\bx_{1:n},\bx_{1:n}; \btheta_{1:n},\btheta_{1:n})
K^{\rm T}( \bx^\dagger_i,\bx_{1:n};\btheta^*_{k}, \btheta_{1:n})  \bigg],
\\
3. \:\:\: &  \mbox{cov} (y^\dagger_i, y^\dagger_{i'}\mid y_{1:n}, \btheta_{1:n}) = \sum_{{k_2}=1}^\infty \sum_{{k_1}=1}^\infty p_{k_1} p_{k_2} \bigg[
K( \bx^\dagger_i,\bx^\dagger_{i'};\btheta^*_{k_1},\btheta^*_{k_2})
\\
&-K( \bx^\dagger_i,\bx_{1:n};\btheta^*_{k_1}, \btheta_{1:n}) K^{-1}(\bx_{1:n},\bx_{1:n}; \btheta_{1:n},\btheta_{1:n})
K^{\rm T}( \bx^\dagger_{i'},\bx_{1:n};\btheta^*_{k_2}, \btheta_{1:n})\bigg].
  \end{aligned}
\end{equation*}
\end{remark}
Here we slightly abused the notation by assuming $K(\bx,\bx';\btheta, \btheta')$ is the matrix (or vector) as formed in \eqref{np_model} between location sets (or points) $\bx$ and $\bx'$, with corresponding parameters $\btheta$ and $\btheta'$.

Lastly, we show an equivalent generative process leading to the same non-stationary GP, which is exploited for efficient computation in the next section.

\begin{remark}
(Computational Form)
 Consider two white-noise Gaussian processes over {$\bw \in [0,\infty)^d$,}
\begin{equation*}
	\begin{aligned}
        \ba(\bw) & \stackrel{indep}{\sim} \No(\bzero,\bI_d), \quad
    \bbb(\bw) \stackrel{indep}{\sim} \No(\bzero,\bI_d),  \:\:\ba(\bw)\text{ and } \bbb(\bw) \text{ are independent,}
	\end{aligned}
\end{equation*}
where $\No(\bzero, \bI_d)$ represents a $d$-variate normal distribution with mean as $\bzero$ and the variance-covariance matrix as the $d$-dimensional identify matrix $\bI_d$.

For $\bw\in \mathbb{R}^d \setminus [0,\infty)^d$, we define
\begin{equation*}
	\begin{aligned}
    \ba(\bw) = \ba(|\bw|), \quad \bbb(\bw)= \left[\prod_{\ell=1}^d\left(\frac{\bw_\ell}{|\bw_\ell|}\right)\right]\bbb(|\bw|)
	\end{aligned}
\end{equation*}
with $|\bw|$ denotes taking absolute value in each sub-dimension $w_1,\ldots, w_d$.

Then the following random variable on $\bx_i\in \mathcal{X}$ specified as
\begin{equation*}
    y(\bx_i)  = \mu(\bx_i) +\int_{\mathbb{R}^d} \exp(j \bx_i^{\rm T} \bw)[g^{1/2}(\bw;\btheta_i)\ba(\bw) /\sqrt{2} + j g^{1/2}(\bw;\btheta_i)\bbb(\bw)/\sqrt{2} ]d\bw + \epsilon_i
\end{equation*}
is a real-valued Gaussian process over $\mathcal{X}$ with the covariance function specified as in \eqref{np_model}, where $\pi(\btheta_i) = \sum_{k=1}^{\infty} p_k \delta_{\btheta^*_k} $ and $\epsilon_i \stackrel{indep}{\sim} \No(0,\sigma^2)$.

\end{remark}

\section{Posterior Computation}\label{estimation}

Based on the property in Remark 4, we now describe an efficient estimation procedure for posterior sampling. The Fourier transform is commonly approximated by the Discrete Fourier Transform (DFT) defined by
\begin{equation*}
	\begin{aligned}
    \frac{1}{|\bW|}\sum_{\bw_j\in \bW} \exp(j\bx^{\rm T}\bw) h(\bw),
	\end{aligned}
\end{equation*}
where $\bW = \times_{\ell=1}^d\{ -\pi+ \frac{1}{m_{\ell}+1}2\pi, -\pi+ \frac{2}{m_{\ell}+1}2\pi,\ldots,
-\pi+ \frac{m_{\ell}}{m_{\ell}+1}2\pi\}$ is the set of equally spaced points in $(-\pi,\pi)^d$, with $|\bW|=\prod_{\ell=1}^d m_{\ell}$, which assumes $h(\bw)\approx 0$ if $\bw$ {$\not \in$} $(-\pi,\pi)^d$. Defining a similar integer set $\bX = \times_{\ell=1}^d\{ 0, 1,\ldots,m_{\ell}\}$, with large enough $m_{\ell}$'s, one can apply scaling and shifting on observed $\bx_{1:n}$ to have $\tilde \bx_{1:n} \in \bX$.

Using $\bQ$ to represent the matrix formed by $\exp(j \bx \bw)$ with $\bx\in \bX$ and $\bw\in \bW$, one convenient result is
\begin{equation*}
	\begin{aligned}
    \bQ\bQ^*=\bQ^*\bQ=\bI,
	\end{aligned}
\end{equation*}
where $\bQ^*$ denotes the transpose conjugate of $\bQ$. We use $\bQ_i $ denote its $i$th row and $\bG(\btheta)$ as the diagonal matrix formed by $g(\bw;\btheta)$. Let $\tilde{y}^{(k)}_i$ denote the augmented observations for each {$\bx_{i}\in \mathcal{X}$}, such that
\begin{equation*}
	\begin{aligned}
    \tilde{y}^{(k)}_i & = y_i, \quad & \text{ if } \btheta_i =\btheta^*_k \text{ and } \bx_i\in \tilde x_{1:n};\\
    \tilde{y}^{(k)}_i & \sim \No( \bQ_i \bG^{1/2}(\btheta_k) (\boldsymbol a + j \boldsymbol  b), \sigma^2), &\text{ otherwise.}
     	\end{aligned}
\end{equation*}

The augmented likelihood based on $\tilde{\bf y}^{(k)}$ is
\begin{equation*}
        \begin{aligned}
        \label{aug_likelihood}
        \mc L(\btheta;\tilde y)= &
  \prod_{k=1}^{k_0}\frac{1}{\sigma^{|\bW|}} \exp  \left( - \frac{1}{2\sigma^2}
\| \bQ^*\tilde{\bf y}^{(k)} - \bG^{1/2}(\btheta_k) (\boldsymbol a + j\boldsymbol b)\|_c^2\right)
\\ &\times \prod_{i=1}^n p_k^ {1(\btheta_i =\btheta^*_k)}\exp  (-\|\boldsymbol a\|^2/2-\|\boldsymbol b\|^2/2),
        \end{aligned}
\end{equation*}
where $\|.\|_c$ is the complex modulus. For convenience, we truncate the number of components to $k_0$ and use $\text{Dir}(\alpha/k_0, \ldots, \alpha/k_0)$ to approximate the Dirichlet process. This augmented likelihood can be rapidly evaluated without matrix inversion. Therefore, it enjoys extremely efficient computation.

 We assign an inverse-gamma prior for $\sigma^2\sim \text{IG}(2,1)$. The posterior can be sampled via MCMC:

\begin{enumerate}
        \item For $i=1,\ldots,n$, update $\btheta_i$ via  multinomial draw from $\{\btheta^{\ast}_1,\ldots,\btheta^{\ast}_{k_0}\}$ based on $$\text{pr}(\btheta_i = \btheta^*_k)=p_k \No(y_i \mid \bQ_i \bG^{1/2}(\btheta_k^*) (\boldsymbol a + j\boldsymbol b), \sigma^2).$$
        \item  Update {the latent variables}  $\tilde{y}^{(k)}_i  \sim \No( \bQ_i G^{1/2}(\btheta_k) (\boldsymbol a + j \boldsymbol b), \sigma^2)$ for those $\bx_i\not\in \tilde \bx_{1:n}$ or $\btheta_i\neq \btheta^*_k$.
        \item For each $\bw\in \bW \cap [0,\pi)^d$, update $a(\bw)$ and $b(\bw)$
\begin{equation*}
	\begin{aligned}
    & a(\bw) \sim \No (\mu_{\ba}, \tau), \quad b(\bw) \sim \No ( \mu_{\bbb}, \tau) \\
    &\tau = [\sum_{k=1}^{k_{0}} g(\bw;\btheta_k)/\sigma^2 +1]^{-1},  \\
    &\mu_{\ba}=\tau \{\sum_{k=1}^{k_{0}} g^{1/2}(\bw;\btheta_k) /\sigma^2 \text{Re}[ \bQ^*(\tilde \by^{(k)})]_{\bw}\},  \quad
    \mu_{\bbb}=\tau\{\sum_{k=1}^{k_{0}} g^{1/2}(\bw;\btheta_k) /\sigma^2 \text{Im}[ \bQ^*(\tilde \by^{(k)})]_{\bw}\},
	\end{aligned}
\end{equation*}
where $\text{Re}$ and $\text{Im}$ denote the real and imaginary parts; subscript $._\bw$ denotes the row corresponding to the index of $\bw$ in $\bW$.
        \item Update $\sigma^2\sim \text{IG}\big (n/2+2, \sum_k \|\tilde \by_k- \bQ \bG^{1/2}(\btheta_k^*) (\boldsymbol a + j \boldsymbol b) \|_c^2/2+1 \big).$
        \item  Update $\btheta^*_k$ for $k=1\ldots k_0$ using Metropolis-Hasting algorithm, via accepting new state $\btheta^{*\prime}$ with the probability
\begin{equation*}
	\begin{aligned}
        \min\bigg[\frac{\mathcal{L}(\btheta^{*\prime})\pi_0(\btheta^{*\prime})h(\btheta^*; \btheta^{*\prime}) }{ \mathcal{L}(\btheta^*)\pi_0(\btheta^*) h(\btheta^{*\prime}; \btheta^{*}) },1\bigg].
	\end{aligned}
\end{equation*}
	    \item Update $[p_1,\ldots,p_{k_0}]\sim \text{Dir}(\alpha/k_0+ \sum_{i=1}^{n} 1({\btheta_i=\btheta_1^*}), \ldots, \alpha/k_0+ \sum_{i=1}^{n} 1({\btheta_i=\btheta_{k_0}^*}))$
        \end{enumerate}
For simplicity, $\btheta^*_k$ is proposed from random walk $h(\btheta^{*\prime}; \btheta^{*}) = \text{Uniform}_{\bTheta} (\btheta^{*}-s,\btheta^{*}+s)$ with truncation to the appropriate domain $\bTheta$; $s$ is tuned during an adaptation period for a good acceptance rate near $0.234$. In the following simulations and data application, the first half of iterations were discarded as burn-inf and the rest were used in posterior inference.

\section{Comparison with Mixture of Independent GPs}\label{iGP}

We now compare our model with the class of models using the mixture of independent GPs (IGP) \citep{rasmussen2002infinite}. Without loss of generality, they can be viewed as a modification based on \eqref{np_model}:
\begin{equation}
\label{IGP_model}
	\begin{aligned}
     y(\cdot)  & \sim \text{GP} \bigg( \mu(\cdot), \tilde K(\cdot, \cdot) \bigg), \\
    \tilde K(\bx_i,\bx_{i'}; \btheta_i,\btheta_{i'}) & =
    \left\{ \begin{array}{llc}
         &  K(\bx_i,\bx_{i'}; \btheta_i,\btheta_{i}),  \quad &\text{ if } \btheta_i=\btheta_{i'};\\
         & 0,\quad & \text{ otherwise,}\\
 \end{array}
 \right.\\
    \pi(\btheta_i) & = \sum_{k=1}^{\infty} p_k \delta_{\btheta^*_k}(\btheta_i),
	\end{aligned}
\end{equation}
where the covariance is $0$ between two components in IGP; whereas it can be non-zero in MSGP. A very useful consequence is that the MSGP covariance reduces the prediction variance.

\begin{remark}
(Prediction Efficiency)
Assuming that two GPs with dependent (as in \eqref{np_model}) or independent components (as in \eqref{IGP_model}) share the same set of values for $p_k$'s and $\btheta_{1:n}$, then the prediction variance for any unobserved $y^\dagger_i$ has
$$\mbox{var}_{MSGP} (y^\dagger_i \mid y_{1:n}, \btheta_{1:n}) \le
\mbox{var}_{IGP} (y^\dagger_i \mid y_{1:n}, \btheta_{1:n}).
$$
\end{remark}

Note that this inequality holds regardless of the sign of $K(\bx_i,\bx_{i'}; \btheta_i,\btheta_{i'})$. Although we obtained this result assuming IGP and MSGP share the same estimate for $\{\btheta^*_k, p_k\}_{k=1}^{\infty}$, we found empirically that the inequality holds well when two models are trained individually.

To illustrate, we generate data $y_i$ from a non-stationary process over location $x_i=1,\ldots,100$ in $\mathbb{R}$, when clearly there are two stationary regions, $R_1=\{x \le 50\}$ and $R_2=\{x> 50\}$ (Figure~\ref{two_part_ns_process}).  We conducted $3$ experiments, by reserving $3$ different testing regions that are within $R_1$, $R_2$  and across $R_1$ and $R_2$. In each experiment, we trained the IGP and MSGP models with squared exponential covariance $\tilde K(x_{i}-x_{{i}^{\prime}};\btheta^*_k)$ on the remaining data. The two models were trained separately without requiring the parameters to be the same.

 The predicted means and variances were obtained for the reserved region.  MSGP shows lower prediction variance for all three regions (Figures~\ref{ns_process_var_g1},\ref{ns_process_var_g2} and \ref{ns_process_var_g12}). In comparing the root-mean-squared error (RMSE) between the prediction mean and the true data, the IGP has RMSE at $3.89, 2.24,4.12$ for three regions, while MSGP has RMSE at lower values as $3.70, 1.85,1.99$.

\begin{figure}[H]
	\centering
	\begin{subfigure}[t]{.9\columnwidth}
		\centering\includegraphics[width=0.5\columnwidth]{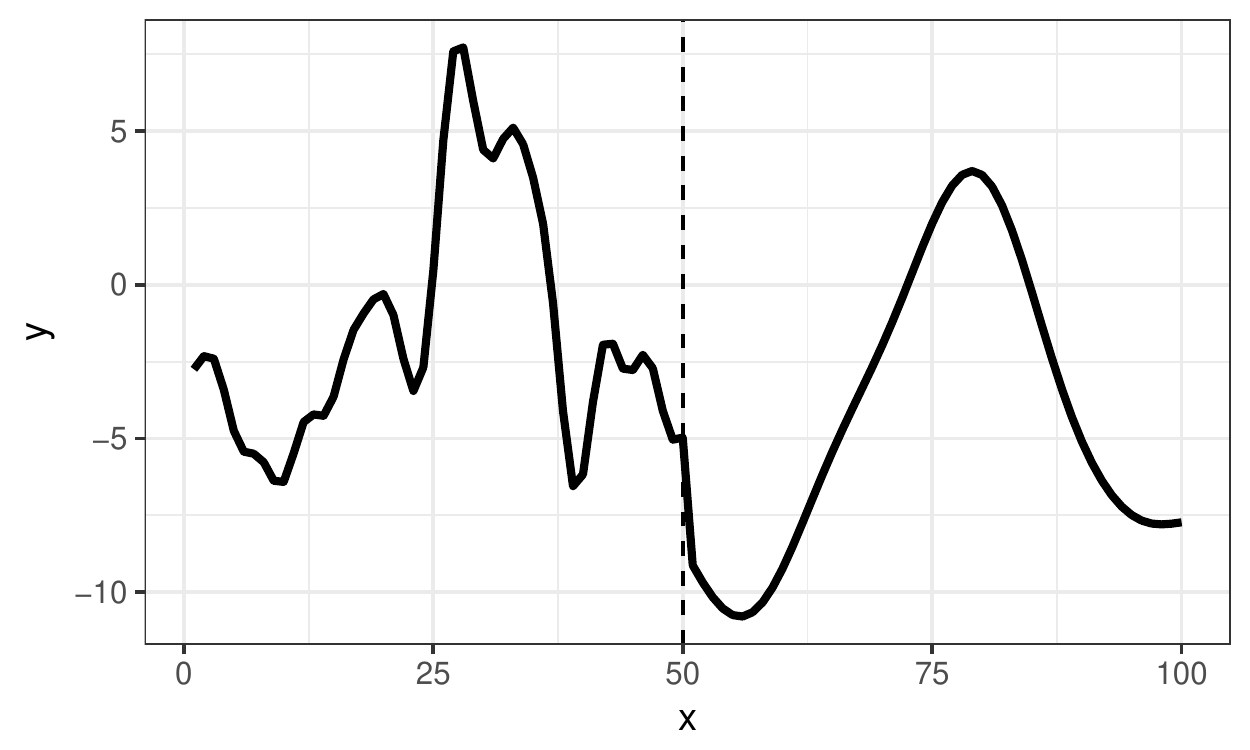}
		\subcaption{Simulated data generated from a non-stationary process. The right part $R_2=\{x> 50\}$ has longer range of autocorrelation than the left $R_1=\{x \le 50\}$.\label{two_part_ns_process}}
	   \end{subfigure}
		\begin{subfigure}[t]{.3\columnwidth}
		\centering\includegraphics[width=1\columnwidth]{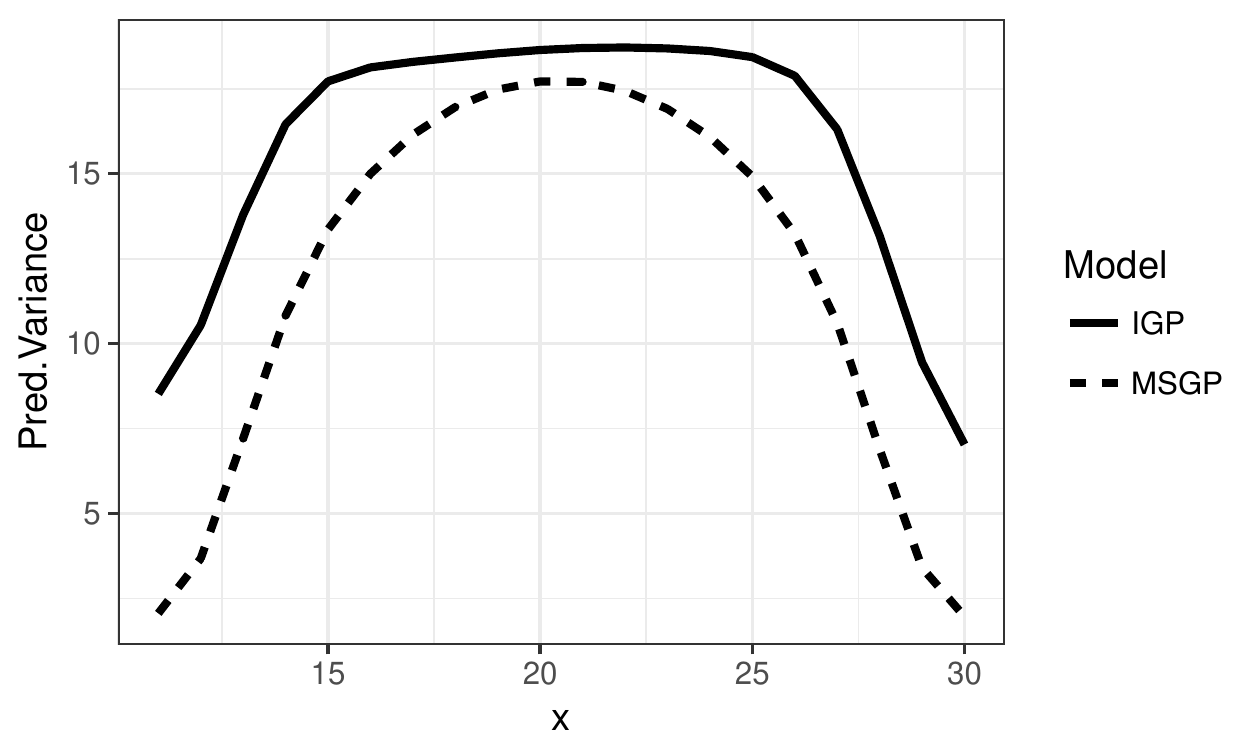}
		\subcaption{Prediction variance for a region $x\in (10,30)$ in the left $R_1$. \label{ns_process_var_g1}}
	\end{subfigure}
	\quad
	\begin{subfigure}[t]{.3\columnwidth}
		\centering\includegraphics[width=1\columnwidth]{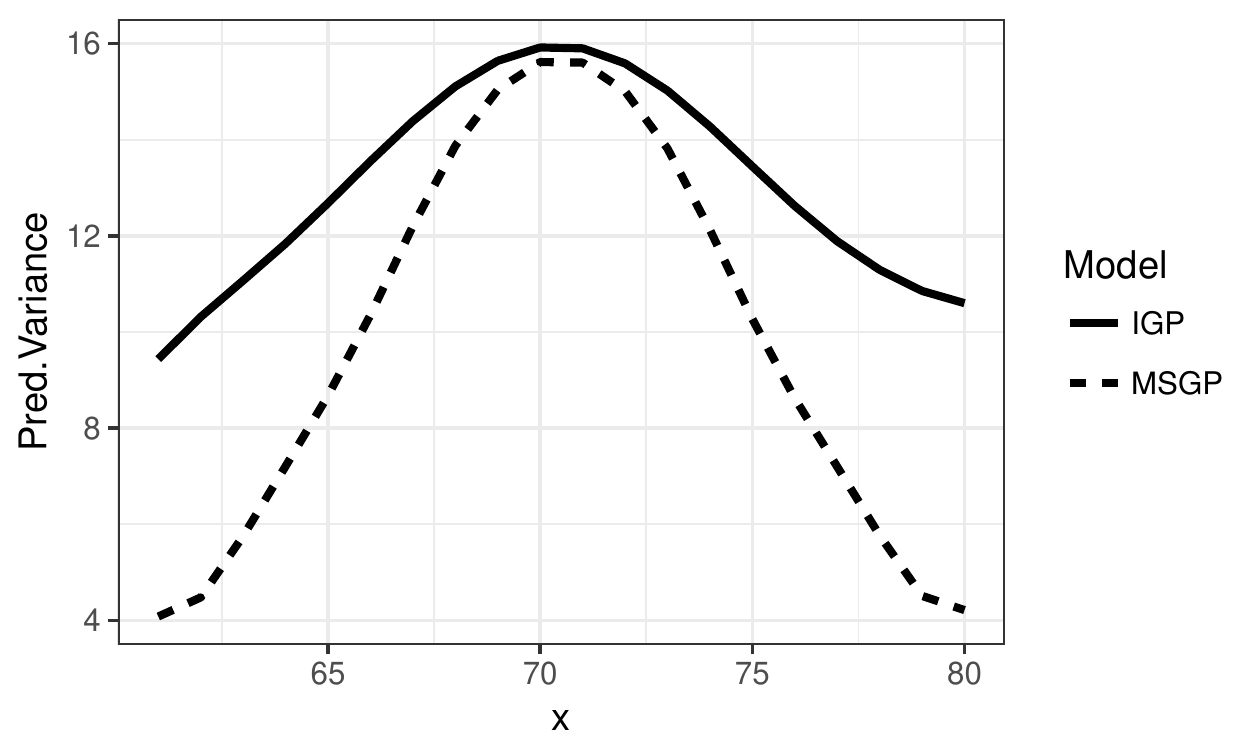}
		\subcaption{Prediction variance for a region $x\in (60,80)$  in the right $R_2$. \label{ns_process_var_g2}}
	\end{subfigure}
			\quad
	\begin{subfigure}[t]{.3\columnwidth}
		\centering\includegraphics[width=1\columnwidth]{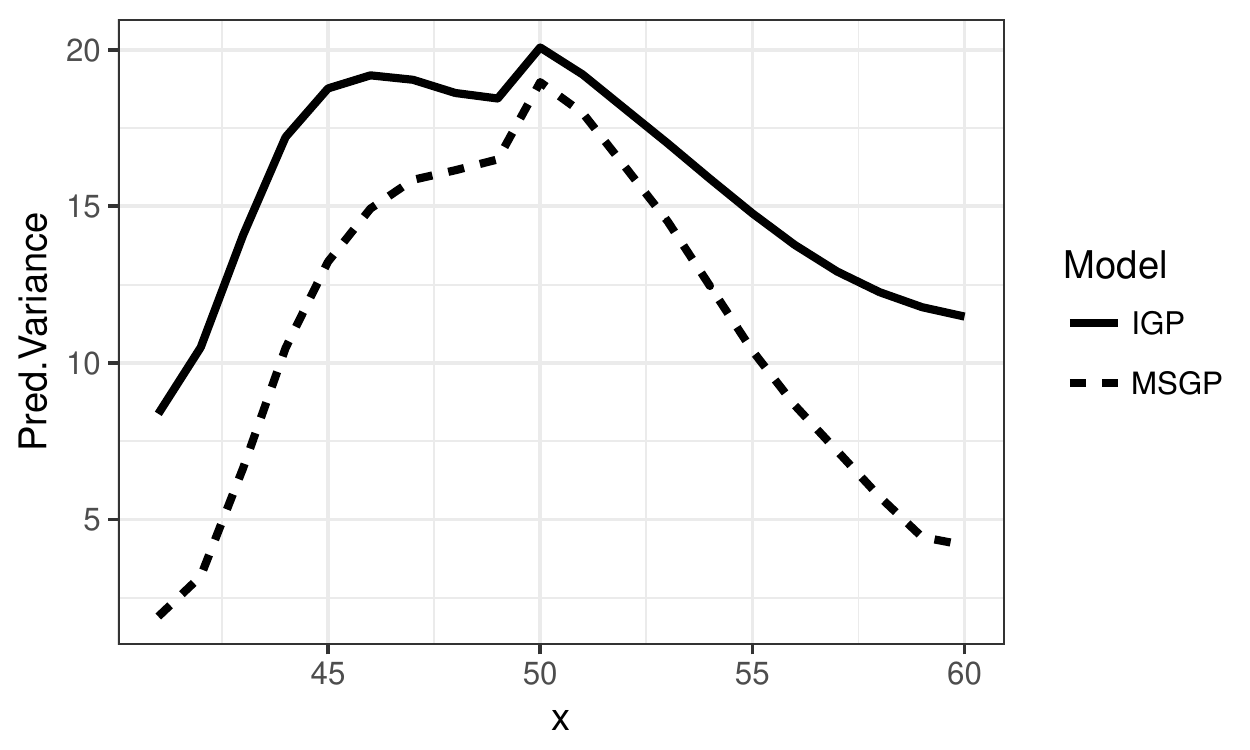}
		\subcaption{Prediction variance for a region $x\in (40,60)$ connecting $R_1$ and $R_2$. \label{ns_process_var_g12}}
	\end{subfigure}
	\caption{Simulation of fitting to a non-stationary data (panel a) with two stationary regions. For all different regions (panel b,c,d), the proposed model (MSGP) has lower prediction variance compared to the mixture of independent GPs (IGP), indicating higher efficiency.}
\end{figure}

\section{Simulation and Data Application}\label{application}

\subsection{Estimation of Stationary Regions in Simulation}

We first illustrate the posterior discreteness of $\btheta_i$ using simulation. To simulate non-stationary surface data, we use the the formulation described by \cite{pintore2004spatially}, which is a locally squared exponential covariance
\begin{equation*}
\begin{aligned}
& K({ \bx_i, \bx_{i'}})= \phi h_{ \bx_i, \bx_{i'}}\exp(-||{ \bx_i- \bx_{i'}}||^2/\theta_{ \bx_i, \bx_{i'}}),
\qquad  \theta_{ \bx_i, \bx_{i'}}=\frac{\beta({ \bx_i})+\beta({ \bx_{i'})}}{2}, \\
& h_{ \bx_i, \bx_{i'}} = \frac{2\beta({ \bx_i})^{1/2}\beta({ \bx_{i'}})^{1/2}}{\beta({ \bx_i})+\beta({ \bx_{i'}})},
\qquad \beta({ \bx_i})=2\rho(\bx_i)^2.
\end{aligned}
\end{equation*}
To assign values, we use a function $\rho(\bx_i)= [\cos(4\pi x_{i1}/100)+2] \exp(x_{i2}/200)$ with $\bx_i =( x_{i1},x_{i2}) \in (0,100)^2$ (Figure~\ref{surface_sim_par}), generating data with varying range of correlation (Figure~\ref{surface_sim_z}). Obviously, this continuous function is complicated and unknown; MSGP's discrete representation can provide a useful approximation.

\begin{figure}[htbp]
  \centering
  \begin{subfigure}[t]{.4\columnwidth}
    \centering\includegraphics[height=2.3in, trim=2cm +1cm 0 +1.3cm]{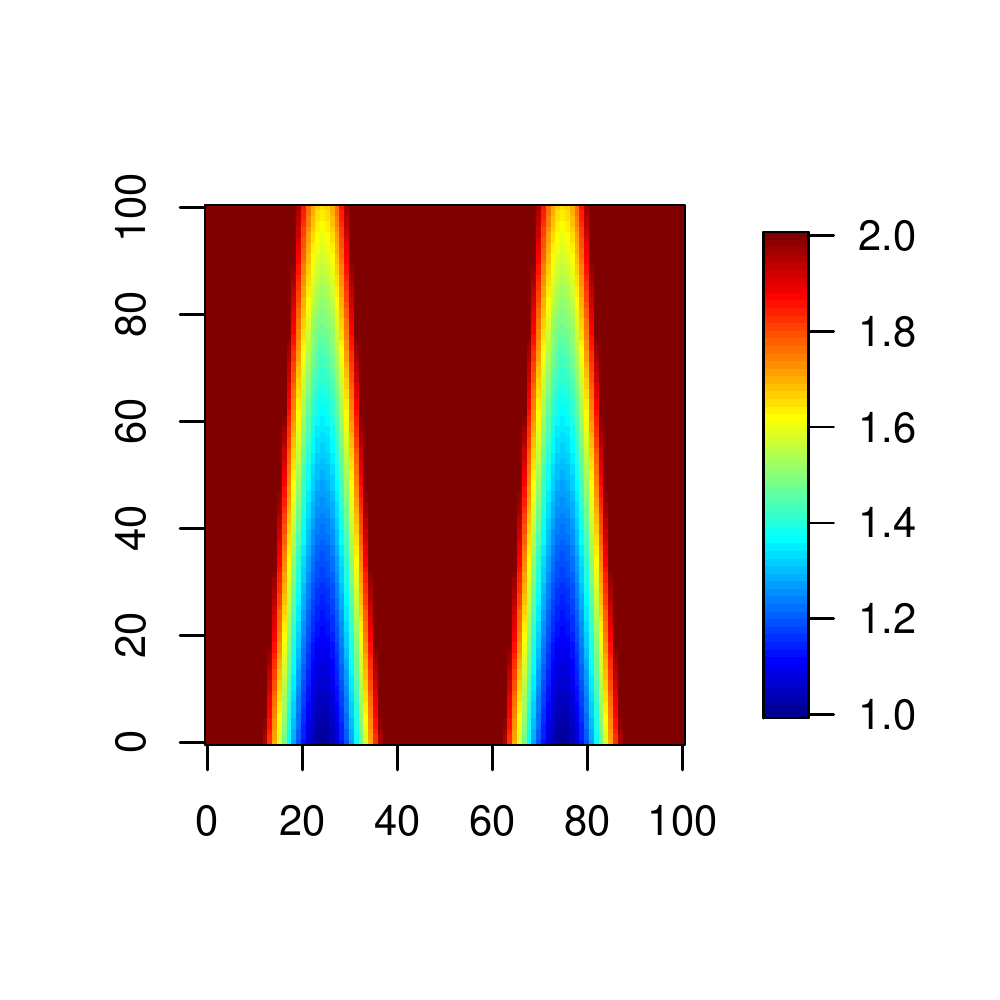}
    \subcaption{Parameters $\theta$ over a 2-D space used to generated a non-stationary surface.\label{surface_sim_par}}
     \end{subfigure}
     \quad
  \begin{subfigure}[t]{.4\columnwidth}
    \centering\includegraphics[height=2.4in]{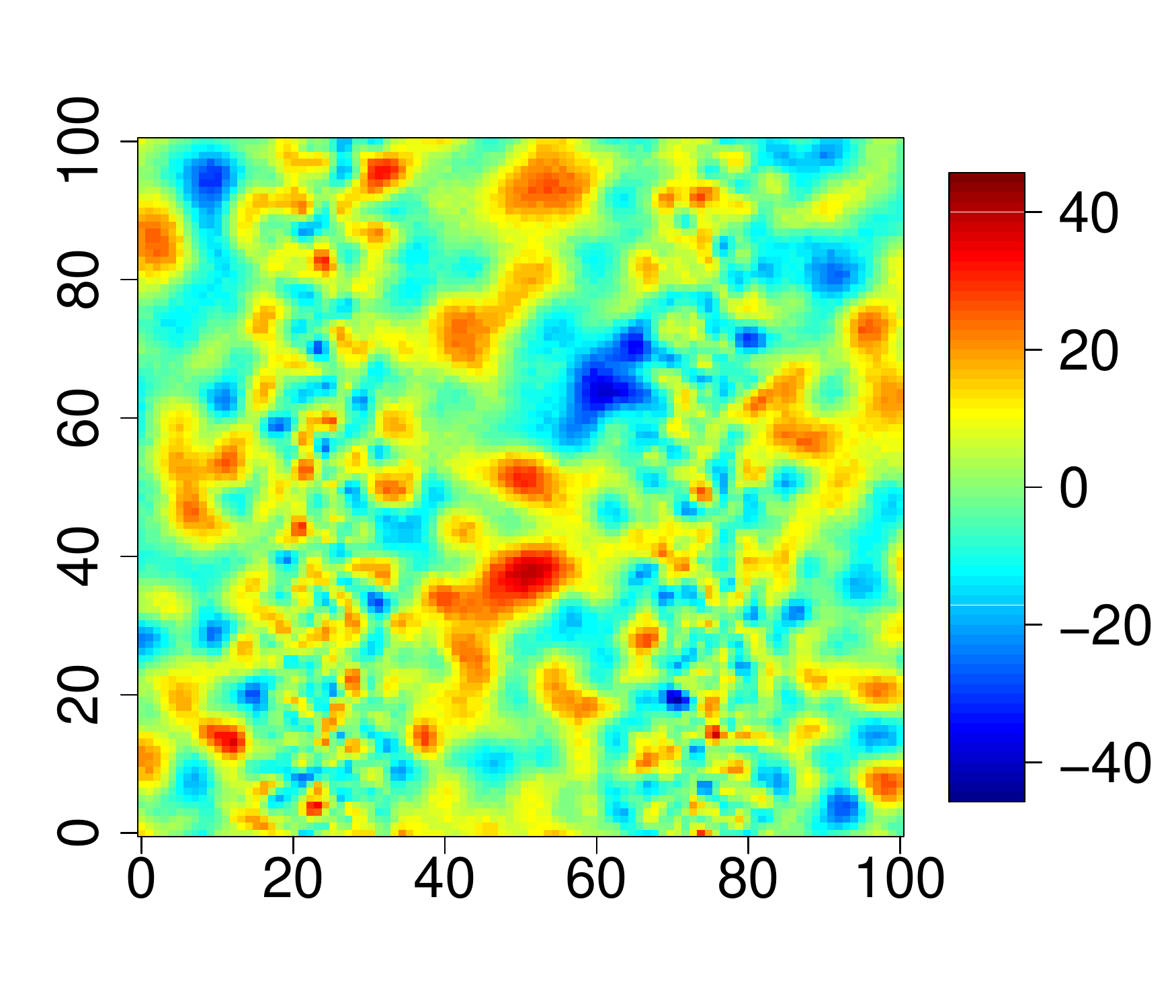}
    \subcaption{Simulated non-stationary surface using $\theta$ over a 2-D space.\label{surface_sim_z}}
     \end{subfigure}
    \begin{subfigure}[t]{.3\columnwidth}
    \centering\includegraphics[width=1\columnwidth]{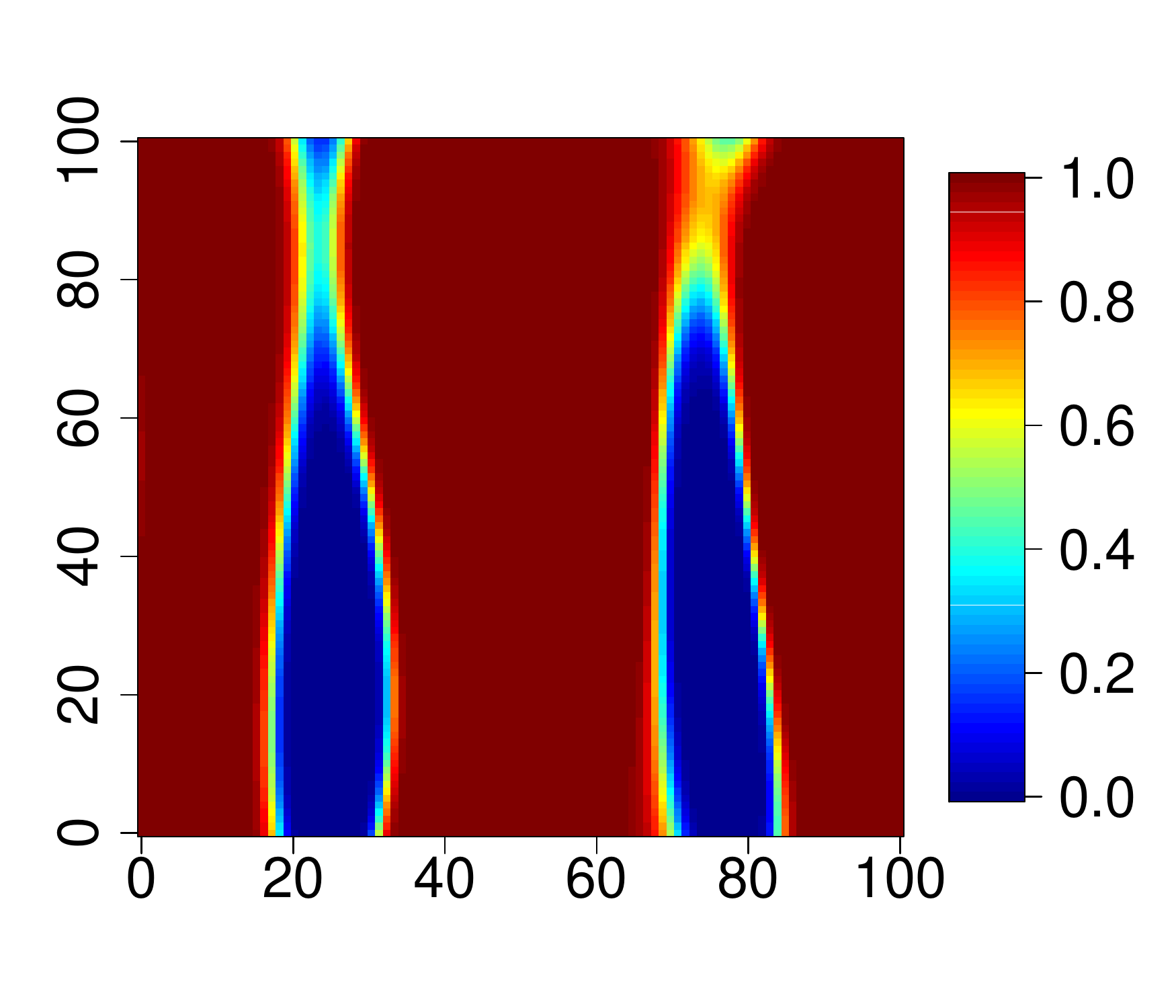}
    \subcaption{Posterior of parameter assigning to first component $\text{pr}(\theta_i=\theta^*_1 \mid y)$\label{sim_w1},  with $\theta^*_1$ mean (std.dev) $1.9 (0.05)$.}
  \end{subfigure}
  \quad
  \begin{subfigure}[t]{.3\columnwidth}
    \centering\includegraphics[width=1\columnwidth]{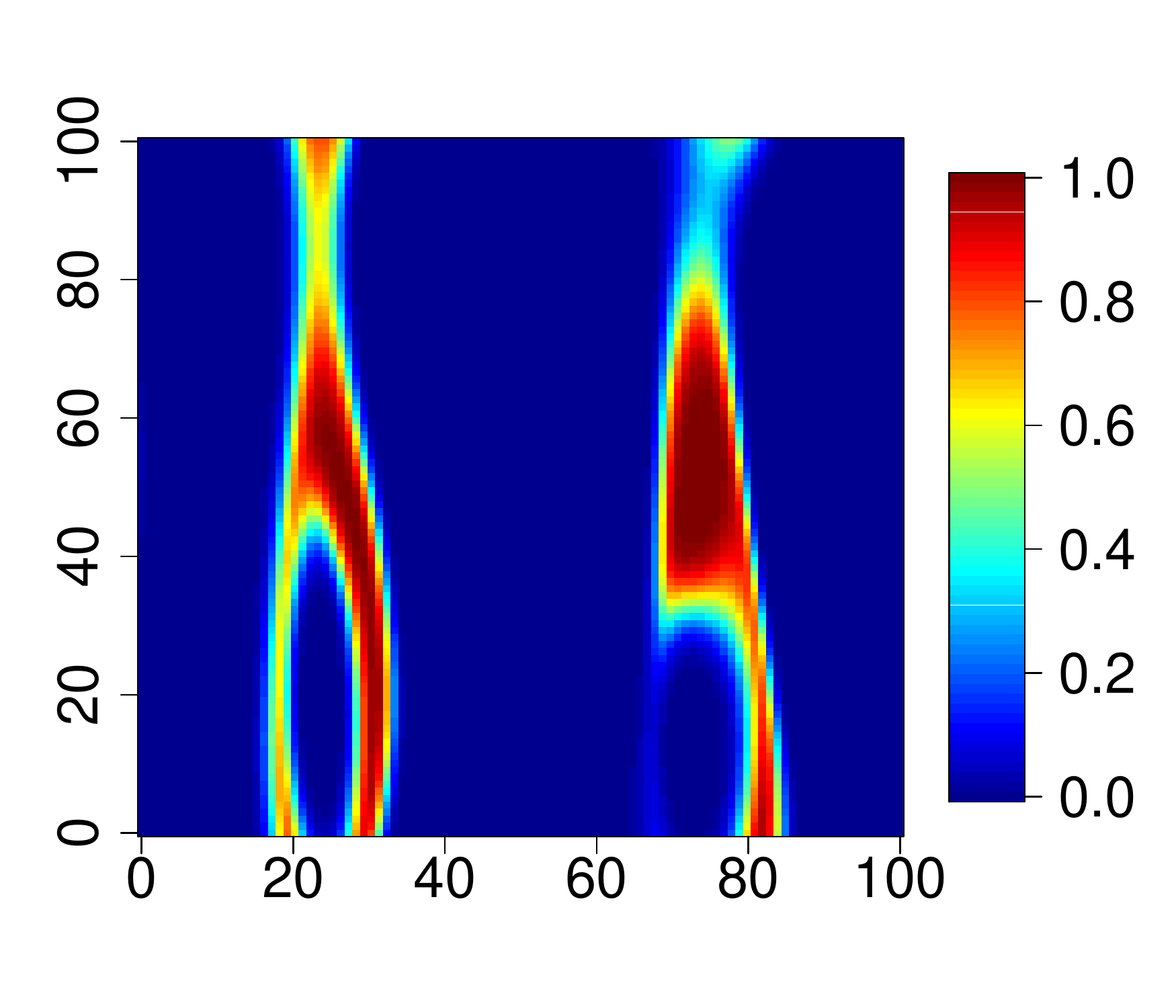}
    \subcaption{Posterior of parameter assigning to second component $\text{pr}(\theta_i=\theta^*_2 \mid y)$\label{sim_w2}, with $\theta^*_2$ mean (std.dev)$1.55 (0.03)$.}
  \end{subfigure}
      \quad
  \begin{subfigure}[t]{.3\columnwidth}
    \centering\includegraphics[width=1\columnwidth]{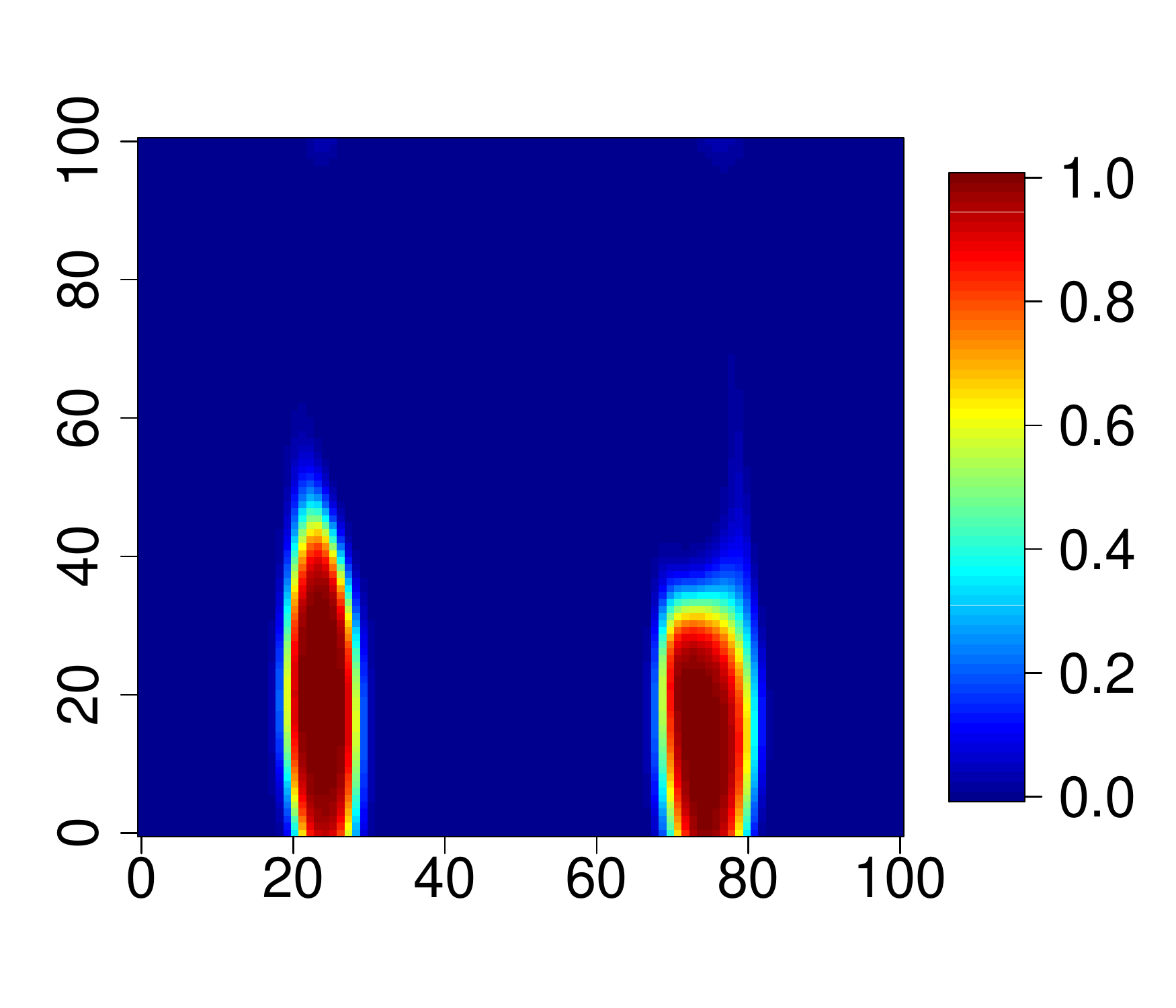}
    \subcaption{Posterior of parameter assigning to third component $\text{pr}(\theta_i=\theta^*_3 \mid y)$\label{sim_w3}, with $\theta^*_3$ mean (std.dev) $1.10 (0.08)$}
  \end{subfigure}
    \begin{subfigure}[t]{.3\columnwidth}
    \centering\includegraphics[width=1\columnwidth]{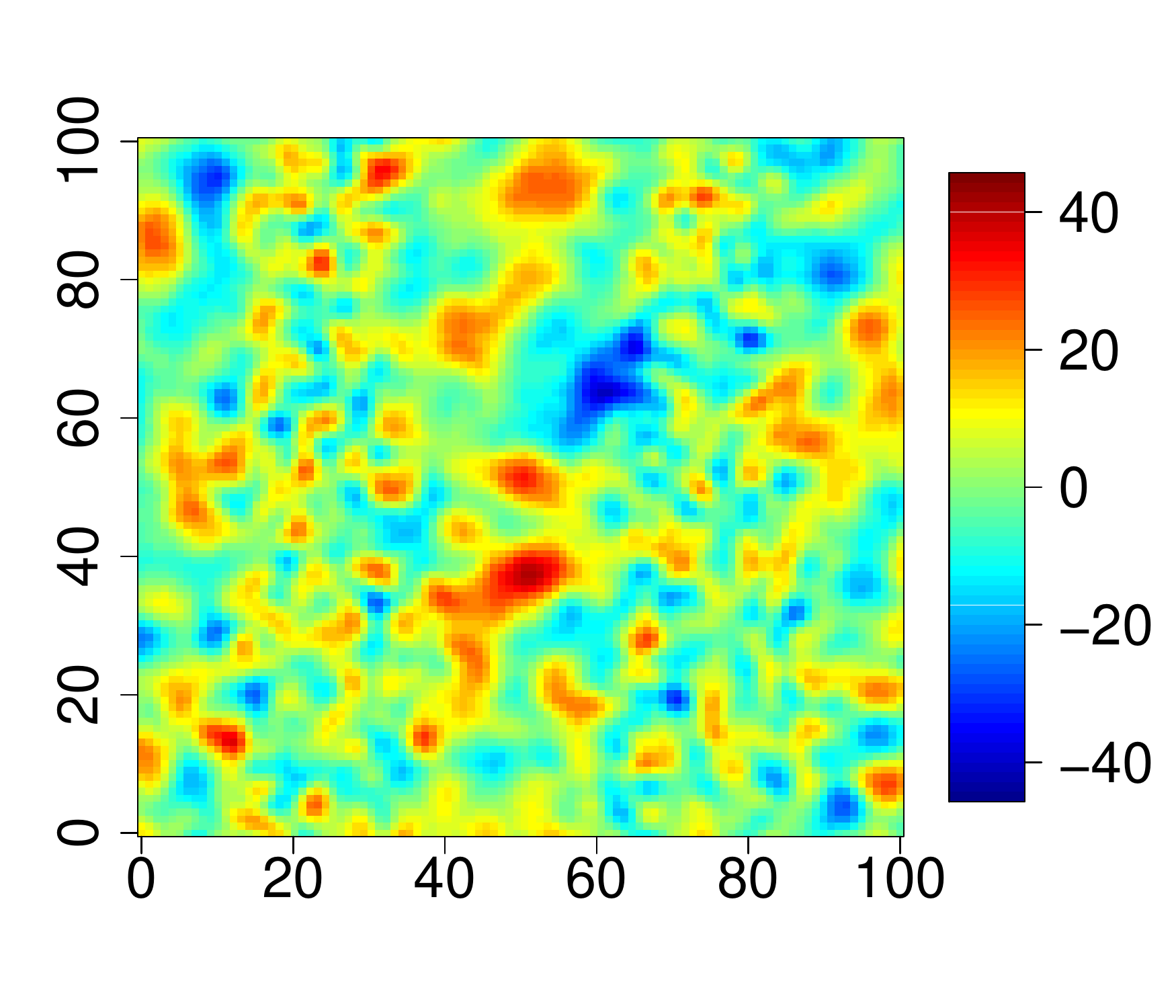}
    \subcaption{First component GP\label{sim_mu1}.}
  \end{subfigure}
  \quad
  \begin{subfigure}[t]{.3\columnwidth}
    \centering\includegraphics[width=1\columnwidth]{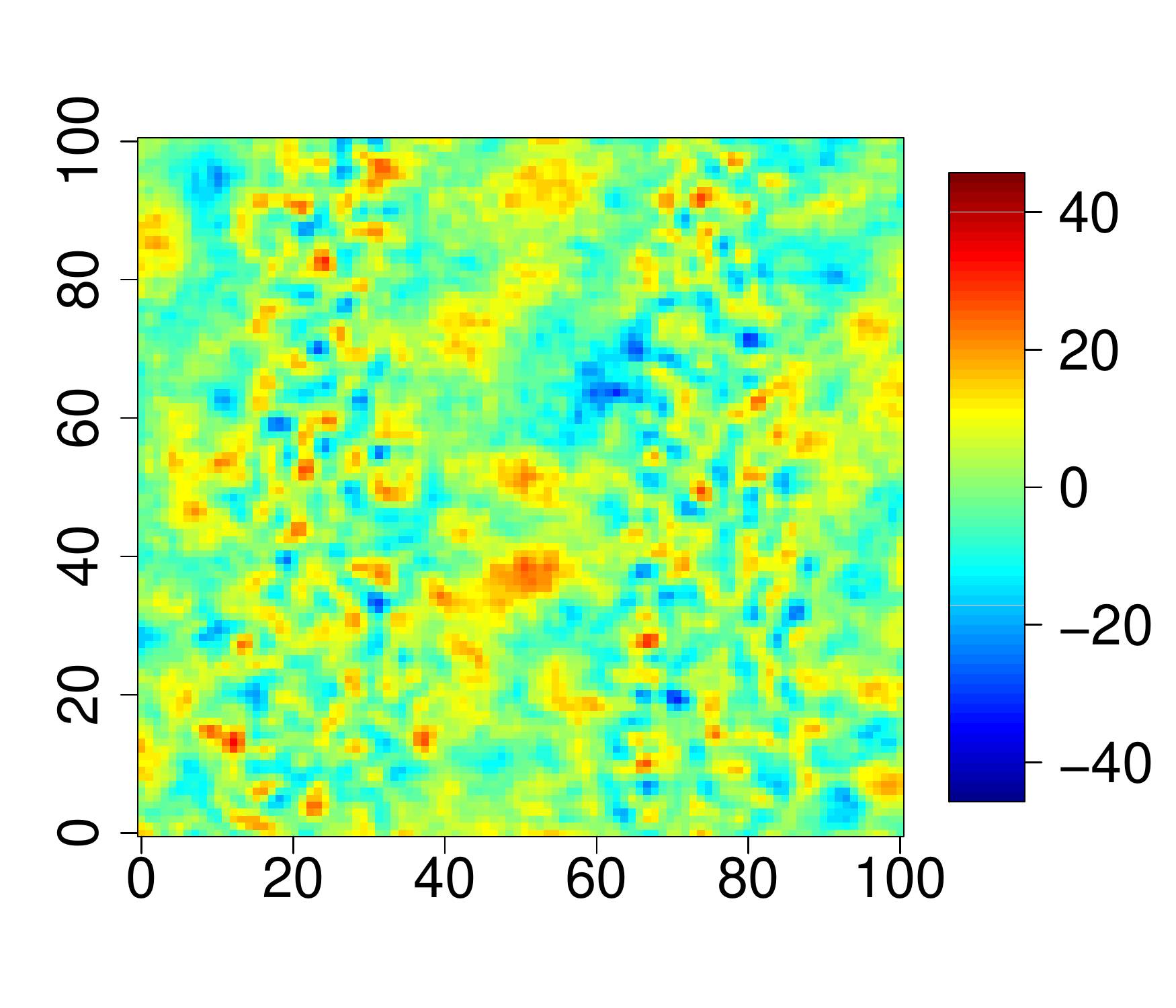}
    \subcaption{Second component GP\label{sim_mu2}.}
  \end{subfigure}
      \quad
  \begin{subfigure}[t]{.3\columnwidth}
    \centering\includegraphics[width=1\columnwidth]{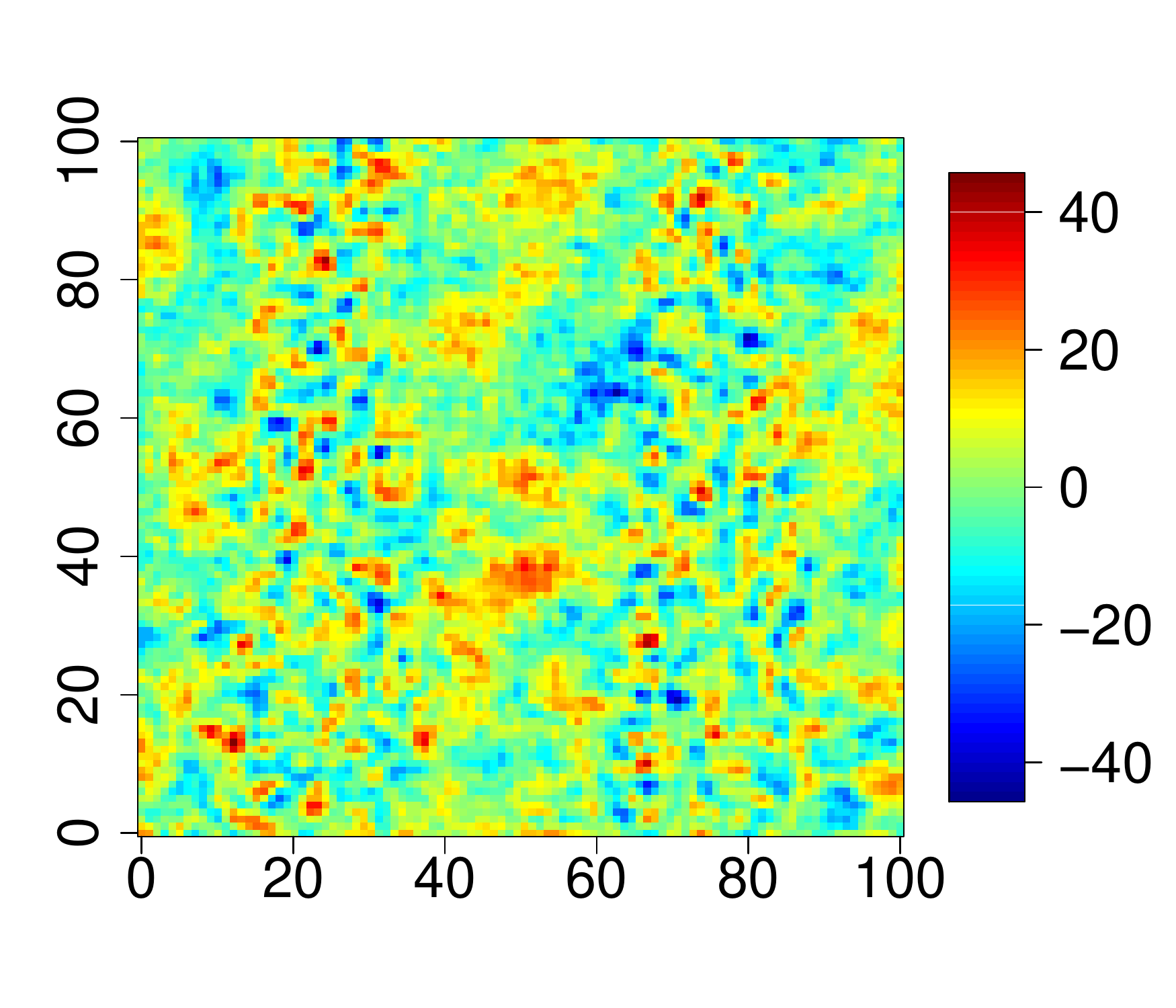}
    \subcaption{Third component GP\label{sim_mu3}.}
  \end{subfigure}
  \caption{Using MSGP to estimate the parameters of a non-stationary GP. Panel (a) shows the ground truth, a hideen continuous function over space; panel (b) shows the generated spatial outcome; panel (c)(d)(e) show the posterior probability of assigning to one of three discrete components; panel (f)(g)(h) shows the dependent GP components.}
\end{figure}

We fitted MSGP models to the data using Dirichlet concentrations $\alpha =1$, $\alpha =0.5$, $\alpha =0.1$ in three experiments. They converged to $6$, $3$ and $3$ dominating components, respectively. For conciseness, we present the $3$-component result here with $\alpha=0.5$. The posterior  $\text{pr}(\theta_i=\theta^*_k)$ for $k=1,2,3$ are visualized in  Figure~\ref{sim_w1}, \ref{sim_w2} and \ref{sim_w3}. This discrete distribution of  the parameter highly resembles the distribution of ground truth (Figures~\ref{surface_sim_par}). We also plot the three component GPs (Figure~\ref{sim_mu1}, \ref{sim_mu2} and \ref{sim_mu3}) $f^*(\bx_i; \btheta^*_k)$ using the mixture of dependent GPs representation. The correlation of those GPs can be clearly observed.

To validate the model, we carried out cross-validation by randomly leaving out $20\%$ of data and computing the RMSE between the predicted and original data. The MSGP model has RMSE at $1.81$, which is very close the oracle model with the ground truth covariance RMSE at $1.78$.

\subsection{Application in Spatial-Temporal Tracking of Temperature}

We now apply the MSGP to a large spatial-temporal dataset. The data are obtained from the North American Regional Climate Change Assessment Program (NARCCAP). We use the surface air temperature in the North America region collected during 1971-2000 \citep{mearns2011north}. The data are simulations from the Weather Research \& Forecasting regional model (WRF) coupled with the Third Generation Coupled Global Climate Model (CGCM3). To exclude seasonal effects, we choose the annual average in each summer, which spans from June 21 to September 21, as the yearly measurement on a $134 \times 109$ dense grid over 30 years. This results in 438,180 data points. We model the temperature $y$ as a non-stationary GP, indexed by $\bx_i=(x_{i1},x_{i2},x_{i3})$ in the 3-dimensional spatial-temporal domain; where $x_{i1}$ denotes the longitude index, $x_{i2}$ the latitude index and $x_{i3}$ the year, respectively.
\begin{equation*}
	\begin{aligned}
        y(\cdot)  & \sim \text{GP}\bigg[\mu(\cdot),  K(\cdot, \cdot) \bigg], \\
        \mu(\bx_i) & = \beta_0 + \beta_{11} x_{i1} + \beta_{12} x_{i2} + \beta_{13} x_{i3}+ \beta_{21} x^2_{i1} + \beta_{22} x^2_{i2} + \beta_{23} x^2_{i3}, \\
        K(\bx_i,\bx_{i'}; \btheta_i,\btheta_{i'}) & = \int_{\mathbb{ R}^d} \exp(j (\bx_{i}-\bx_{i'})^{\rm T} \bw) g^{1/2}(\bw; \btheta_{i})g^{1/2}(\bw; \btheta_{i'}) d\bw + \sigma^2 1(i=i'),\\
\pi(\btheta_i) & = \sum_{k=1}^{\infty} p_k \delta_{\btheta^*_k}(\btheta_i),
	\end{aligned}
\end{equation*}
where we use a polynomial function to capture the naive trend in $\mu(\bx_i)$. To encourage parsimonious result, we choose $\alpha=0.1$ as the Dirichlet concentration.

Here the component specific parameters is denoted as $\btheta^* = (\phi, \rho_1,\rho_2,\rho_3,c_1,c_2)$. Since $g$ depends $\tilde K (\bx_{i}-\bx_{{i}^{'}};\btheta^*)$, we consider the  spatial-temporal covariance including time-space interaction terms, proposed by \cite{cressie1999classes}
\begin{equation*}
	\begin{aligned}
	  & \tilde K[(x_{i1},x_{i2},x_{i3} )-(x_{i'1},x_{i'2},x_{i'3} ); \btheta]  =\phi \exp\left(- \frac{|x_{i1}-x_{i'1}|^2}{2\rho_1^2} - \frac{|x_{i2}-x_{i'2}|^2}{2\rho_2^2} \right)
	   \\ & \cdot \exp \left(- \frac{|x_{i3}-x_{i'3}|}{\rho_3}\right )
		\cdot \exp\left(- \frac{|x_{i1}-x_{i'1}|^2|x_{i3}-x_{i'3}|}{c_1}- \frac{|x_{i2}-x_{i'2}|^2 |x_{i3}-x_{i'3}|}{c_2}\right),
	\end{aligned}
\end{equation*}
where {$c_1 > 0$ and $c_2>0$}. Then
 \begin{equation*}
  g((w_1,w_2,w_3);\btheta)  =  \int_{\mathbb{ R}^3} \exp(-j \bx^T \bw) f^*(\bx; \btheta)d\bx.
 \end{equation*}Note that even though $g((w_1,w_2,w_3);\btheta)$ does not have a closed-form, it can be numerically evaluated via fast Fourier transform. We applied the model on the data and ran MCMC sampling for 30,000 steps and use the last 15,000 steps as the posterior sample. The process took about 9 hours on a quad-core laptop.

The model converged to 4 dominating components. We omit the components that are too small in size. Table~\ref{Tab:4_Components_NS}  lists the parameter estimates. Figure~\ref{Fig:4_Components_NS} plots the dependent GP component mean $f^*(\bx_i;\btheta_k^*)$, and the probability of assigning data to it $\text{pr}(\btheta_i= \btheta^*_k \mid y)$, over $k=1, 2, 3, 4$ . The component means have different range parameters, exhibiting varying strength of autocorrelation over space. The temporal parameters $\rho_3$ are relatively large and imply strong temporal correlation. As compared between Panels (a) and (c) in Figure~\ref{Fig:4_Components_NS}, a subtle evolution of temperature pattern can been observed from 1990 to 1995. As compared between Panels (b) and (d), the temporal change in component weights is hardly noticeable. This suggests that the temperature pattern mainly changes in time by changing each stationary component, with their weights fixed. Interestingly, component 2 shows some space-time interaction with small $c_1$ and $c_2$.

\begin{table}[ht]
	\begin{center}
		\begin{tabular}{ r  | r r r r}
			\hline
			& Component 1&	Component 2 & Component 3 & Component 4\\
			\hline
			\hline
			$\rho_1$ & 12.17 (0.13) & 4.11 (0.07) & 8.26  (0.09) & 8.47  (0.13) \\
			$\rho_2$  &10.89 (0.08) & 4.87 (0.07) & 8.50  (0.08) & 9.39 (0.19)\\
			$\rho_3$  & 16.43  (0.58) & 45.61  (2.33) & 29.04  (1.48) & 23.07  (1.85)\\
			$c_1$  & 6347 (451) & 61.72 (2.01) & 2400 (145)& 1738 (204)\\
			$c_2$  &4615 (239) &   73.38 (1.21) &2396 (146) & 2083 (94)\\
			$\phi$  &6.71  (0.39)  & 3.30  (0.07)& 10.89  (0.71)& 11.63  (0.56)\\
						\hline
			$\sigma^2$  & \multicolumn{4}{c}{1.82 (0.05)}\\
			\hline
		\end{tabular}
	\end{center}
	\caption{The parameter estimates (mean and standard deviation) in MSGP spatial-temporal model with non-separable covariance.}
	\label{Tab:4_Components_NS}
\end{table}

\begin{figure}[hp]
	\centering
	\begin{subfigure}[t]{.9\columnwidth}
		\centering\includegraphics[width=1\columnwidth]{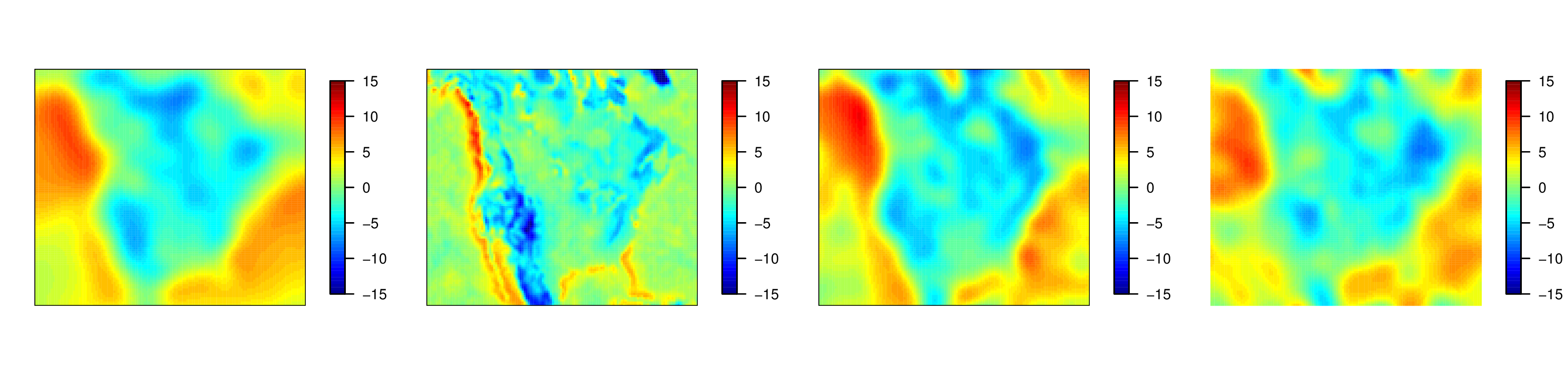}
		\subcaption{
	The component stationary GP	 mean $f^*(\bx_i;\btheta^*_k)$  for $k=1,2,3, 4$ Year 1990}
	\end{subfigure}
	\begin{subfigure}[t]{.9\columnwidth}
		\centering\includegraphics[width=1\columnwidth]{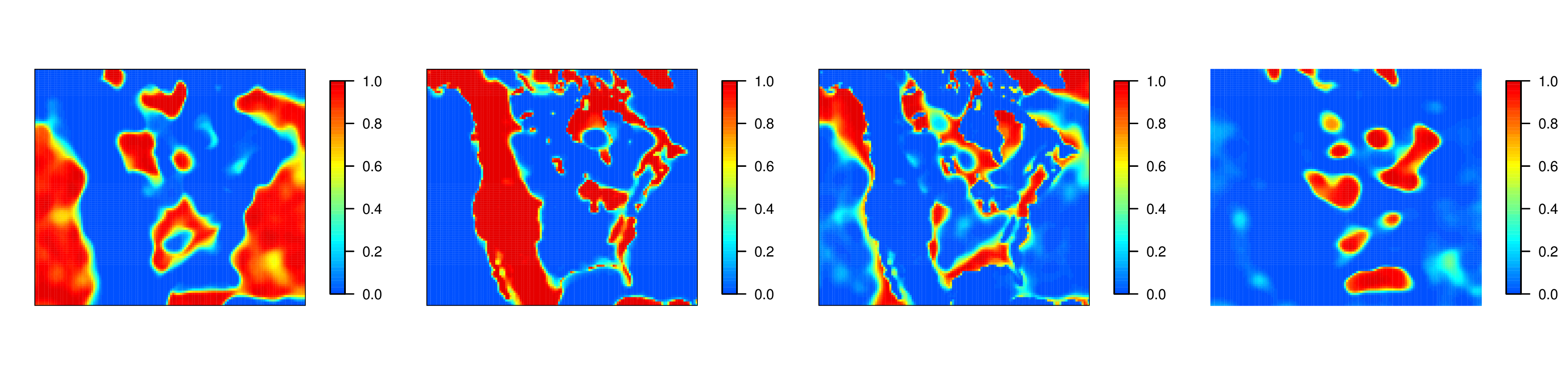}
		\subcaption{Correponding $\text{pr}(\btheta_i= \btheta^*_k \mid y)$ for $k=1,2,3, 4$ Year 1990}
	\end{subfigure}
	\begin{subfigure}[t]{.9\columnwidth}
		\centering\includegraphics[width=1\columnwidth]{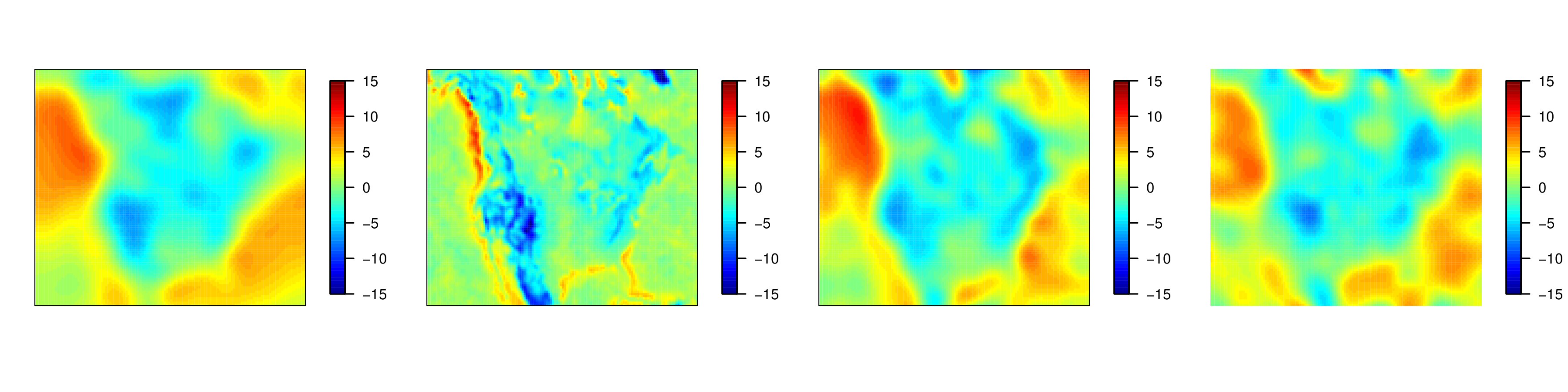}
		\subcaption{
	The component stationary GP	 mean $f^*(\bx_i;\btheta^*_k)$  for $k=1,2,3, 4$ Year 1995}
	\end{subfigure}
	\begin{subfigure}[t]{.9\columnwidth}
		\centering\includegraphics[width=1\columnwidth]{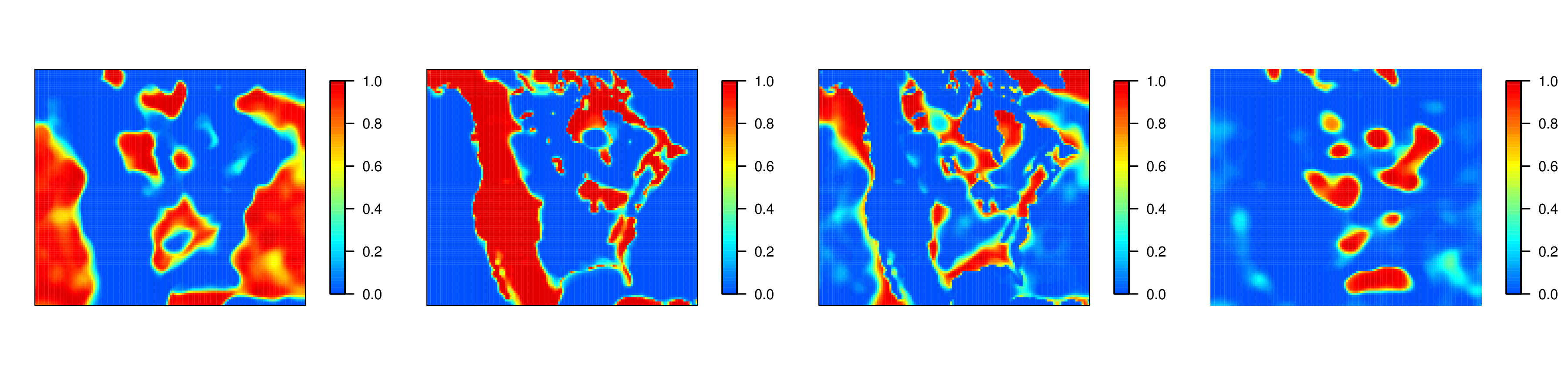}
		\subcaption{Corresponding $\text{pr}(\btheta_i= \btheta^*_k \mid y)$ for $k=1,2,3, 4$ Year 1995}
	\end{subfigure}
	\caption{Four stationary components estimated by the time-space interactive model MSGP.}
	\label{Fig:4_Components_NS}
\end{figure}

For benchmarking, we carried out cross-validation by masking all the observations in 1996-2000 and predicted the outcome. We also compared with two baseline models, the MSGP model with separable covariance with $c_1$ and $c_2$ at infinity, and the IGP model with non-separable covariance. The prediction errors in RMSE are shown in Table~\ref{Tab:Forecasting_error}. With the space-time interaction, the non-separable MSGP model shows a slight improvement over the separable MSGP  model in terms of performance. {Both MSGP models perform much better compared to IGP with non-separable covariance.}  For illustration, we plot the prediction map of temperature data of the year 1996 in Figure~\ref{Fig:prediction_plot}.  For a comprehensive comparison, we also tested the treed GP model  \citep{gramacy2012bayesian} on a smaller set of the data, with model averaging, we found its prediction accuracy is similar to MSGP with separable covariance, although the prediction uncertainty is larger than MSGP (details listed in {Table S1 of the Supplementary Materials}).

\begin{table}[ht]
	\begin{center}
		\begin{tabular}{l || l l  l l l   }\hline
			\multicolumn{6}{c}{Average Prediction Uncertainty $(\sum_i \mbox{var} (y)^{\dagger}_i/n)^{1/2}$} \\
			\hline
			Model   &    1996 & 1997 & 1998 & 1999 & 2000\\
			\hline
			MSGP with Separable Covariance &  1.88 & 1.96 & 1.91 & 2.54 & 3.07\\
			\hline
			IGP with Non-Separable Covariance  &  3.46 & 3.93 & 3.67 & 4.23 & 4.67\\
			\hline
			MSGP with Non-Separable Covariance  &  1.95 & 1.25 & 1.54 & 1.74 & 2.15\\
		\hline
			\multicolumn{6}{c}{Prediction RMSE $(\sum_i (y^{\dagger}_i -{\mathbb{E} y^{\dagger}_i})^2/n)^{1/2}$} \\
			\hline
			Model   &    1996 & 1997 & 1998 & 1999 & 2000\\
			\hline
			MSGP with Separable Covariance & 1.70 & 1.80 & 1.85 & 2.39 & 2.91\\
			\hline
			IGP with Non-Separable Covariance  &  3.21 & 3.51 & 3.62 & 3.90 & 4.27 \\
			\hline
			MSGP with Non-Separable Covariance  &  1.54 & 1.12 & 1.33 & 1.50 & 1.65 \\
			\hline
		\end{tabular}
	\end{center}
	\caption{Prediction uncertainty and accuracy of three models in 1996-2000.}
	\label{Tab:Forecasting_error}
\end{table}

\begin{figure}[ht]
	\begin{subfigure}[t]{.32\columnwidth}
		\centering
		\includegraphics[width=1\columnwidth]{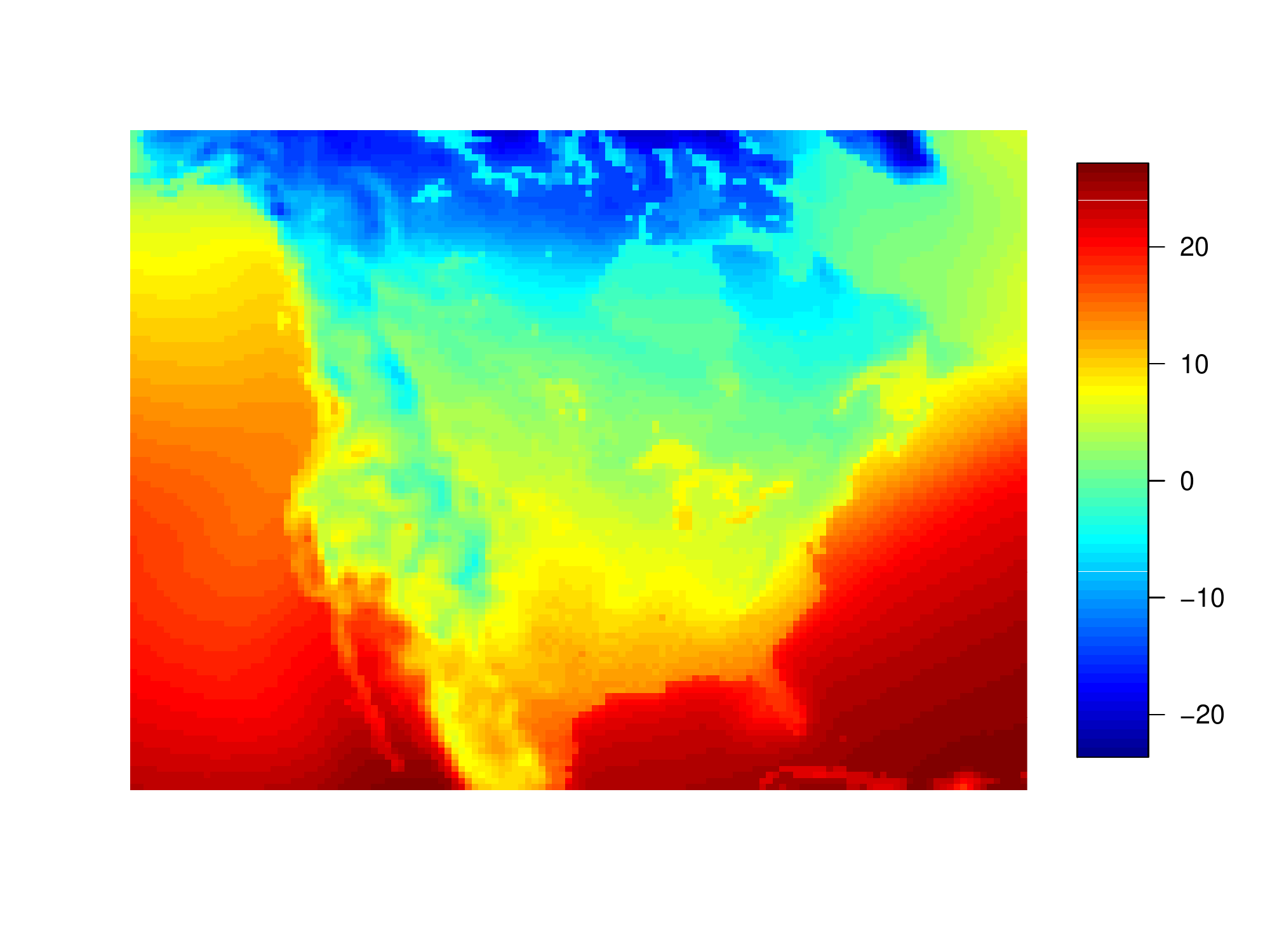}
		\subcaption{The temperature data in 1996 in the original data set. The unit is in Celsius.}
	\end{subfigure} \vrule height -1ex\
	\begin{subfigure}[t]{.32\columnwidth}
		\centering	\includegraphics[width=1\columnwidth]{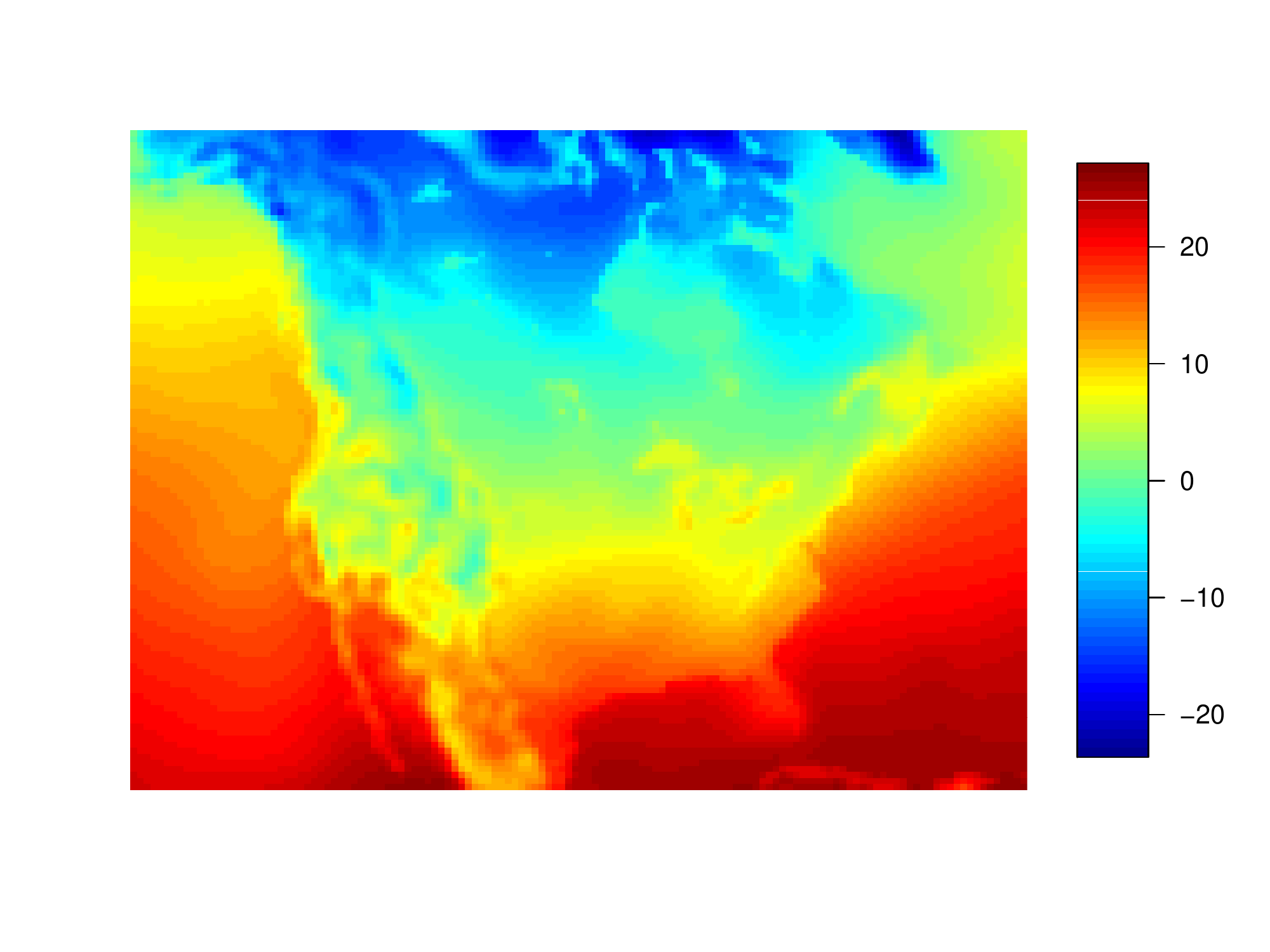}
		\subcaption{The predicted mean of MSGP using non-separable covariance.}
	\end{subfigure} \vrule height -1ex\
	\begin{subfigure}[t]{.32\columnwidth}
		\centering\includegraphics[width=1\columnwidth]{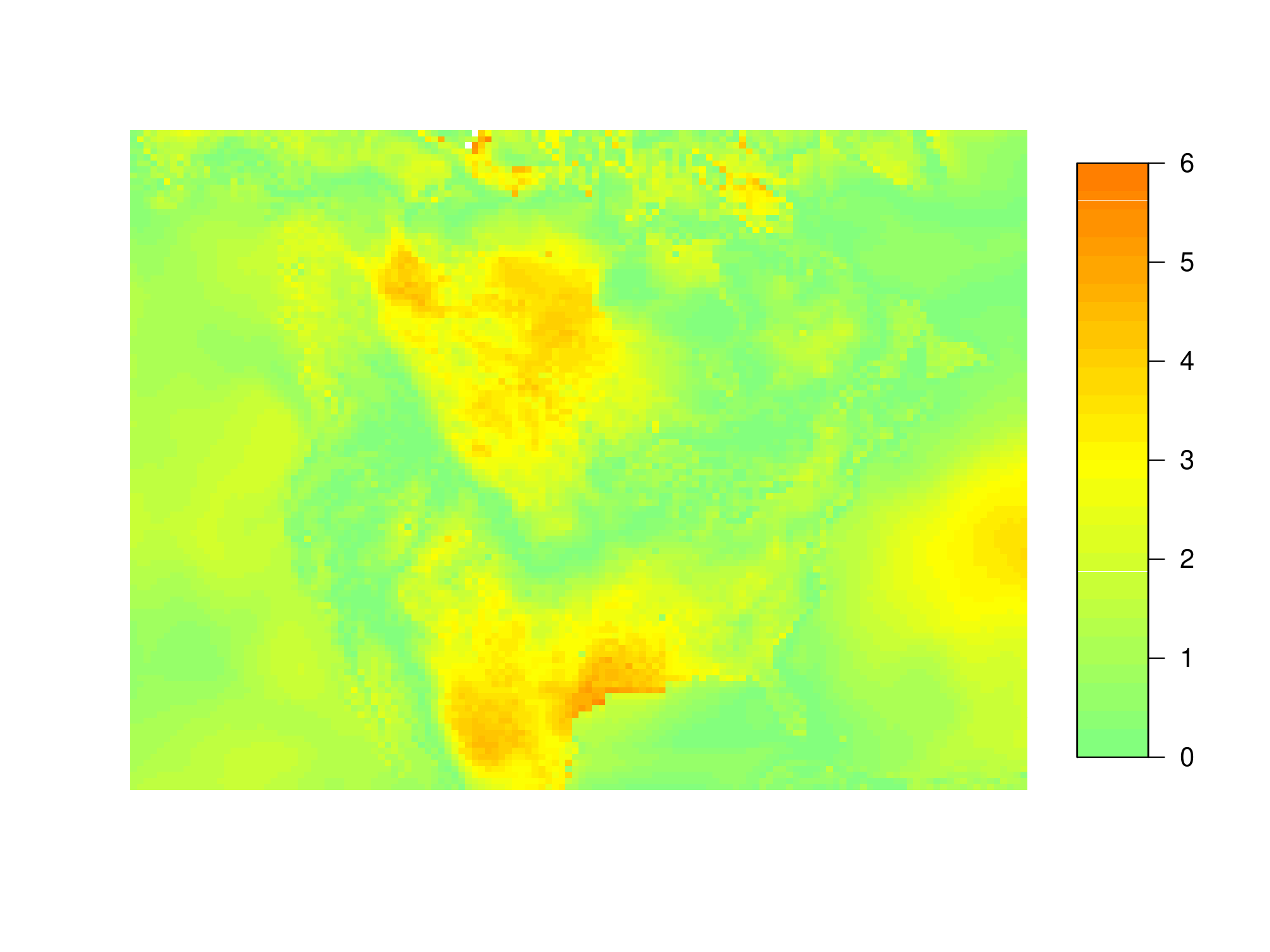}
		\subcaption{The absolute deviation between the real and predicted values.}
	\end{subfigure}
	\caption{Comparison of the original temperatures in the data set and the prediction generated by the space-time interactive model using MSGP.}
	\label{Fig:prediction_plot}
\end{figure}

\section{Discussion}\label{discussion}

In this article, we provide a tractable way to construct a non-stationary GP covariance, with a potentially infinite number of parameters. Exploiting the posterior discreteness yields a substantial parameter reduction in this flexible model. We did not require the component assignment to be spatially correlated, as done in the latent stick-breaking process \citep{rodriguez2010latent} or TreedGP \citep{gramacy2012bayesian}. We expect that further gain could be achieved when this is included. Compared to the mixture of independent GPs, we observed better performance of MSGP in terms of prediction. This is likely due to the data under examination being correlated across regions. One can imagine that for data with distinct spatial patterns, such as precipitation in forest versus desert, the independence assumption would be more appropriate. Therefore, one interesting direction is to develop an encompassing framework to probabilistically switch off the between-component covariance, if strongly indicated by the data.


\bibliographystyle{chicago}

\bibliography{reference}

\section*{Supplementary Materials}

\subsection*{Proof of Remark 1: Komolgorov existence}
\begin{proof}
    It is trivial that $L(y_{1:n};\btheta_{1:n}) \prod_i \pi(\btheta_i =\btheta^*_{k_i})$ is exchangeable in the index. We now prove the second property. By Fubini's theorem:

\begin{equation*}
	\begin{aligned}
 &  \int_{\mathbb{R}} \int_{F_n}\ldots
 \int_{F_1}\bigg\{ \sum_{k_{n+1}\in\{1,\ldots,\infty\}} \sum_{k_n\in\{1,\ldots,\infty\}}\ldots\sum_{k_1\in\{1,\ldots,\infty\}} \\
 & \big [\mc L(y_{1:n};\btheta_{1:n}) \prod_i \pi(\btheta_i =\btheta^*_{k_i})\big]  \bigg\} dy_1 dy_2\ldots dy_n dy_{n+1}\\
= & \int_{F_n}\ldots
 \int_{F_1}\bigg\{ \sum_{k_n\in\{1,\ldots,\infty\}}\ldots\sum_{k_1\in\{1,\ldots,\infty\}}\\
& \bigg( \sum_{k_{n+1}\in\{1,\ldots,\infty\}} \int_{\mathbb{R}} 
   \big [\mc L(\btheta_{1:n+1};y_{1:n+1}) \pi(\btheta_{n+1} =\btheta^*_{k_{n+1}}) \big] dy_{n+1} \bigg) 
   \\&
   \prod_{i=1}^n \pi(\btheta_i =\btheta^*_{k_i})\bigg\} dy_1 dy_2\ldots dy_{n}
   \\	
 = & \int_{F_n}\ldots
 \int_{F_1}\bigg\{ \sum_{k_n\in\{1,\ldots,\infty\}}\ldots\sum_{k_1\in\{1,\ldots,\infty\}}\\
& \mc L(y_{1:n};\btheta_{1:n}) \bigg( \sum_{k_{n+1}\in\{1,\ldots,\infty\}} \pi(\btheta_{n+1} =\btheta^*_{k_{n+1}}) \big] \bigg) \prod_{i=1}^n \pi(\btheta_i =\btheta^*_{k_i})\bigg\} dy_1 dy_2\ldots dy_{n}
   \\
    = & \int_{F_n}\ldots
 \int_{F_1}\bigg\{ \sum_{k_n\in\{1,\ldots,\infty\}}\ldots\sum_{k_1\in\{1,\ldots,\infty\}} \big [\mc L(y_{1:n};\btheta_{1:n})\prod_{i=1}^n \pi(\btheta_i =\btheta^*_{k_i}) \big ]\bigg\} dy_1 dy_2\ldots dy_{n}
   \\	
      \end{aligned}
\end{equation*}
where the second equality is due to the marginalization of Gaussian distribution and third one is due to $\sum_{k=1}^{\infty} p_k=1$.
\end{proof}

\subsection*{Proof of Remark 4: Computational Form}

\begin{proof}
    Verifying the Gaussian assumption is trivial. We prove the real-valuedness of $y$.
    
Note for any $w'_l$ equal to $w$ except for $1$-sub-dimension, that is
$$w'_l= (w_1,w_2,\ldots, -w_l, \ldots,w_{d-1}, w_d).$$
For easier notations, we use

\begin{equation*}
	\begin{aligned}
    A(w) = g^{1/2}(w;\btheta)a(w)/\sqrt{2}, \qquad
    B(w) =  g^{1/2}(w;\btheta)b(w)/\sqrt{2}
	\end{aligned}
\end{equation*}

\begin{equation*}
	\begin{aligned}
    &\exp(j x w)[A(w) + jB(w)] 
    \\ = &     \exp(\sum_{k\in 1,\ldots,d , k\neq l}^d j x_k w_k)  \exp(j x_l w_l)[A(w) + jB(w)] \\ 
    = & \exp(\sum_{k\in 1,\ldots,d , k\neq l}^d j x_k w_k)   (\cos(x_lw_l) + j\sin(x_lw_l))[A(w) + jB(w)]\\
    = & \exp(\sum_{k\in 1,\ldots,d , k\neq l}^d j x_k w_k)  \bigg[ \cos(x_lw_l)A(w) - \sin(x_l w_l)B(w) + j\sin(x_lw_l)A(w) 
     \\& + j\cos(x_lw_l)B(w) \bigg]\\
    = & \exp(\sum_{k\in 1,\ldots,d , k\neq l}^d j x_k w_k)  \bigg[ \cos(-x_lw_l)A(w'_l) - \sin(-x_l w_l)B(w'_l)  \\& - j\sin(-x_lw_l)A(w'_l) -j\cos(-x_lw_l)B(w'_l) \bigg],
	\end{aligned}
\end{equation*}
which means all the imaginary parts are canceled when summing over the $2^d$ orthants

\begin{equation*}
	\begin{aligned}
       & \int_{\mathbb{R}^d} \exp(j x_i w)[A(w) + jB(w)]dw  
=    & 2^d  \int_{(0,\infty)^d}  \text{Re}\bigg( \exp(j x_i w)[A(w) + jB(w)]\bigg) dw,
	\end{aligned}
\end{equation*}
which is real-valued.

Note that,

\begin{equation*}
	\begin{aligned}
 \int_{\mathbb{R}^d} \exp(j x_i w)[A(w) + jB(w)]dw  =
  \int_{\mathbb{R}^d} \exp(-j x_i w)[A(w) - jB(w)]dw
	\end{aligned}
\end{equation*}

The covariance between two location $x_i$ and $x_{i'}$, for $i\neq i'$
    
\begin{equation*}
	\begin{aligned}
    \text{cov}(y_i, y_{i'})   = & \text{cov}\bigg \{
         \int_{\mathbb{R}^d} \exp(j x_i w)[g^{1/2}(w;\btheta_i)a(w)/\sqrt{2} + j g^{1/2}(w;\btheta_i) b(w)/\sqrt{2}]dw, \\
         &
         \int_{\mathbb{R}^d} \exp(- j x_i w)[g^{1/2}(w;\btheta_{i'})a(w)/\sqrt{2} - j g^{1/2}(w;\btheta_{i'}) b(w)/\sqrt{2}]dw
         \bigg \} \\
    = &   \int_{\mathbb{R}^d} \exp(j (x_i-x_{i'}) w)g^{1/2}(w;\btheta_i) g^{1/2}(w;\btheta_{i'})/2 \text{Cov}[a(w)]dw \\
    &- j^2   \int_{\mathbb{R}^d} \exp(j (x_i-x_{i'}) w)g^{1/2}(w;\btheta_i) g^{1/2}(w;\btheta_{i'})/2\text{Cov}[b(w)]dw \\
    = &   \int_{\mathbb{R}^d} \exp(j (x_i-x_{i'}) w)g^{1/2}(w;\btheta_i) g^{1/2}(w;\btheta_{i'})]dw,
    	\end{aligned}
\end{equation*}
which completes the proof.
\end{proof}

\subsection*{Proof of Remark 5: Prediction Efficiency}

\begin{proof}
The two variances are defined as,
$$
\begin{aligned}
 \mbox{var}_{IGP} (y^\dagger_i \mid y_{1:n}, \btheta_{1:n}) & =  \sum_{{k}=1}^\infty p_{k} \bigg[
\tilde K( x^\dagger_i,x^\dagger_i;\btheta^*_{(k)},\btheta^*_{(k)})
\\
&-\tilde  K( x^\dagger_i,x_{1:n};\btheta^*_{k}, \btheta_{1:n}) \tilde K^{-1}(x_{1:n},x_{1:n}; \btheta_{1:n})
\tilde K^T( x^\dagger_i,x_{1:n};\btheta^*_{k}, \btheta_{1:n})  \bigg]
\end{aligned}
$$

$$
\begin{aligned}
\mbox{var}_{MSGP} (y^\dagger_i \mid y_{1:n}, \btheta_{1:n}) & =  \sum_{{k}=1}^\infty p_{k} \bigg[
K( x^\dagger_i,x^\dagger_i;\btheta^*_{(k)},\btheta^*_{(k)})
\\
&-K( x^\dagger_i,x_{1:n};\btheta^*_{k}, \btheta_{1:n}) K^{-1}(x_{1:n},x_{1:n}; \btheta_{1:n})
K^T( x^\dagger_i,x_{1:n};\btheta^*_{k}, \btheta_{1:n})  \bigg]
\end{aligned}
$$

Taking the difference, since $\tilde K( x^\dagger_i,x^\dagger_i;\btheta^*_{(k)},\btheta^*_{(k)})
=K( x^\dagger_i,x^\dagger_i;\btheta^*_{(k)},\btheta^*_{(k)})
$

\begin{equation*}
	\begin{aligned}
    & \mbox{var}_{IGP} (y^\dagger_i \mid y_{1:n}, \btheta_{1:n}) - 
    \mbox{var}_{MSGP} (y^\dagger_i \mid y_{1:n}, \btheta_{1:n}) \\
    = & \sum_{{k}=1}^\infty p_{k} \bigg[
K( x^\dagger_i,x_{1:n};\btheta^*_{k}, \btheta_{1:n}) K^{-1}(x_{1:n},x_{1:n}; \btheta_{1:n})
K^T( x^\dagger_i,x_{1:n};\btheta^*_{k}, \btheta_{1:n}) \\
&- \tilde K( x^\dagger_i,x_{1:n};\btheta^*_{k}, \btheta_{1:n})  \tilde  K^{-1}(x_{1:n},x_{1:n}; \btheta_{1:n})
\tilde K^T( x^\dagger_i,x_{1:n};\btheta^*_{k}, \btheta_{1:n})
 \bigg]
	\end{aligned}
\end{equation*}

We now prove the difference inside the bracket is non-negative for any $k$. We will start with only two mixture components, then extend to infinite components.

Focusing on $k=1$, with two components, one can write the difference in terms of matrix form:

\begin{equation*}
	\begin{aligned}
& \begin{bmatrix}
    B_1'&
    B_2' 
\end{bmatrix}
\begin{bmatrix}
    \Sigma_{11} & \Sigma_{12} \\
    \Sigma_{12}' & \Sigma_{22} \\
\end{bmatrix}^{-1}
\begin{bmatrix}
    B_1\\
    B_2 
\end{bmatrix}
-
\begin{bmatrix}
    B_1'&
    0' 
\end{bmatrix}
\begin{bmatrix}
    \Sigma_{11} & O \\
    O' & \Sigma_{22} \\
\end{bmatrix}^{-1}
\begin{bmatrix}
    B_1\\
    0 
\end{bmatrix} \\
= &
\begin{bmatrix}
    B_1'&
    B_2' 
\end{bmatrix}
\begin{bmatrix}
    \Omega_{11} & \Omega_{12} \\
    \Omega_{12}' & \Omega_{22} \\
\end{bmatrix}
\begin{bmatrix}
    B_1\\
    B_2 
\end{bmatrix}
- B_1'   \Sigma_{11}^{-1} 
    B_1
	\end{aligned}
\end{equation*}
where $B_k$ and $\Sigma_{kk}$ correspond to the sub-vector and sub-matrix formed by \\ $K( x^\dagger_i,x_{1:n};\btheta^*_{1}, \btheta_{1:n}=\btheta^*_{k})$ and $K(x_{1:n},x_{1:n}; \btheta_{1:n}=\btheta^*_{k}, \btheta_{1:n}=\btheta^*_{k})$, $\Sigma_{12}$ is the one by $K(x_{1:n},x_{1:n}; \btheta_{1:n}=\btheta^*_{1}, \btheta_{1:n}=\btheta^*_{2})$; $\Omega_{.}$ is the corresponding matrix blocks in the inverse.

Now using block matrix inversion rule $\Sigma_{11} = (\Omega_{11} - \Omega_{12}\Omega_{22}^{-1} \Omega'_{12})^{-1}$, we have the difference equal to:

\begin{equation*}
	\begin{aligned}
& B_2'\Omega_{22} B_2 +  B_1'\Omega_{12} B_2 + B_2'\Omega'_{12} B_1 + B_1' \Omega_{12}\Omega_{22}^{-1} \Omega'_{12}B_1 \\
= & [\Omega_{22}^{1/2}B_2 +  \Omega_{22}^{-1/2}\Omega_{12}'B_1]'[\Omega_{22}^{1/2}B_2 +  \Omega_{22}^{-1/2}\Omega_{12}'B_1]
	\end{aligned}
\end{equation*}
where $\Omega_{22}^{1/2'}\Omega_{22}^{1/2}=\Omega_{22}$ and $\Omega_{22}^{-1/2'} \Omega_{22}^{1/2} = I$. This quadratic term is non-negative.

Now assuming $h$ components, we have the difference,

\begin{equation*}
	\begin{aligned}
& \begin{bmatrix}
    B_1'&
    B_2' & \cdots & B_h'
\end{bmatrix}
\begin{bmatrix}
    \Sigma_{11} & \Sigma_{12} &\cdots & \Sigma_{1h} \\
    \Sigma_{12}' & \Sigma_{22} &\cdots & \Sigma_{2h} \\
    \cdots & \cdots & \cdots & \cdots \\
    \Sigma_{1h}' & \Sigma'_{2h} &\cdots & \Sigma_{hh} \\
\end{bmatrix}^{-1}
\begin{bmatrix}
    B_1\\
    B_2 \\
    \cdots\\
    B_h
\end{bmatrix}
 \\& -
 \begin{bmatrix}
    B_1'&
    0' & \cdots & 0'
\end{bmatrix}
\begin{bmatrix}
    \Sigma_{11} & O &\cdots & O \\
    O' & \Sigma_{22} &\cdots & O \\
    \cdots & \cdots & \cdots & \cdots \\
    O' & O &\cdots & \Sigma_{hh} \\
\end{bmatrix}^{-1}
\begin{bmatrix}
    B_1\\
    0 \\
    \cdots\\
    0
\end{bmatrix},
	\end{aligned}
\end{equation*}
which is obviously also non-negative by replacing $B_2$ in the two component case by $[B_2 ,\cdots, B_h]$, and changing the corresponding block matrices.

This result applies to $k=2,\ldots,h$, thus 
$$     \mbox{var}_{IGP} (y^\dagger_i \mid y_{1:n}, \btheta_{1:n}) - 
    \mbox{var}_{MSGP} (y^\dagger_i \mid y_{1:n}, \btheta_{1:n}) \ge 0
$$

\end{proof}

\subsection*{Comparison of Prediction Performance with Other Models}
\begin{table}[ht]
	\begin{center}
				\footnotesize	
		\begin{tabular}{l || l l  }
			\multicolumn{3}{c}{Average Prediction Uncertainty $(\sum_i \mbox{var} y^{\dagger}_i/n)^{1/2}$} \\
			\hline
			Model    \\
			\hline
			MSGP with Separable Covariance &  1.75 \\
			\hline
			IGP with Separable Covariance  &  4.01 \\
			\hline
			TreedGP  &  3.20 \\
		\hline
			\multicolumn{3}{c}{Prediction RMSE $(\sum_i (y^{\dagger}_i -{\mathbb{E} y^{\dagger}_i})^2/n)^{1/2}$ } \\
			\hline
			Model   \\
			\hline
			MSGP with Separable Covariance &  1.55\\
			\hline
			IGP with Separable Covariance  &  3.85\\
			\hline
			TreedGP  &  1.54 \\
			\hline
		\end{tabular}
	\end{center}
	\caption{Prediction uncertainty and accuracy in comparing MSGP with IGP and TreedGP using $80\%-20\%$ split cross-validation using the data from 1996.}
\end{table}

\end{document}